\documentclass{article}

\PassOptionsToPackage{numbers, compress}{natbib}
\usepackage[preprint]{neurips_2026}
\usepackage[utf8]{inputenc}
\usepackage[T1]{fontenc}
\usepackage{hyperref}
\usepackage{url}
\usepackage{booktabs}
\usepackage{array}
\usepackage{multirow}
\usepackage{amsfonts}
\usepackage{nicefrac}
\usepackage{microtype}
\usepackage{xcolor}
\usepackage{graphicx}
\usepackage{subcaption} % Support subfigure environment (side-by-side subfigures)
\usepackage{amsmath}
\usepackage{enumitem} % Support enumerate/itemize options
\usepackage{etoc}
\usepackage{fvextra}
\DefineVerbatimEnvironment{verbatim}{Verbatim}{breaklines=true,breakanywhere=true,fontsize=\small}
\graphicspath{{assets/}}

\setlist{leftmargin=1.5em,itemsep=0pt,parsep=0pt,topsep=1pt,partopsep=0pt}
\AtBeginDocument{%
    \setlength{\abovedisplayskip}{6pt plus 2pt minus 2pt}%
    \setlength{\belowdisplayskip}{6pt plus 2pt minus 2pt}%
    \setlength{\abovedisplayshortskip}{4pt plus 2pt minus 2pt}%
    \setlength{\belowdisplayshortskip}{4pt plus 2pt minus 2pt}%
}

\makeatletter
\newcommand{\appendixonlytableofcontents}{\@starttoc{apt}}
\newcommand{\startappendixonlytoc}{%
    \setcounter{tocdepth}{2}%
    \let\slime@orig@addcontentsline\addcontentsline
    \renewcommand{\addcontentsline}[3]{%
        \slime@orig@addcontentsline{##1}{##2}{##3}%
        \def\slime@tocfile{##1}%
        \def\slime@tocname{toc}%
        \ifx\slime@tocfile\slime@tocname
            \slime@orig@addcontentsline{apt}{##2}{##3}%
        \fi
    }%
}
\makeatother

\definecolor{takeawaybg}{gray}{0.93}
\definecolor{takeawayrule}{gray}{0.45}
\newsavebox{\takeawaybox}
\newenvironment{takeaway}[1]{%
    \par\vspace{-1pt}%
    \setlength{\fboxsep}{2.5pt}%
    \begin{lrbox}{\takeawaybox}%
    \begin{minipage}{\dimexpr\linewidth-2\fboxsep\relax}%
    \small
    {\color{takeawayrule}\rule[-0.4ex]{1.5pt}{1.1em}}~\textbf{#1}\par\vspace{0pt}%
}{%
    \end{minipage}%
    \end{lrbox}%
    \noindent\colorbox{takeawaybg}{\usebox{\takeawaybox}}\par\vspace{-1pt}%
}

\title{Signal Reshaping for GRPO in Weak-Feedback Agentic Code Repair}
\author{%
  Jia Li$^{1}$ \quad
  Yuxin Su$^{2}$ \quad
  Ting Peng$^{3}$ \quad
  Hailiang Huang$^{3}$ \quad
  Yuetang Deng$^{3}$ \quad
  Michael R. Lyu$^{1}$ \\
  $^{1}$The Chinese University of Hong Kong \quad
  $^{2}$Sun Yat-sen University \quad
  $^{3}$Tencent \\
  \texttt{linsayli@link.cuhk.edu.hk, suyx35@mail.sysu.edu.cn, lyu@cse.cuhk.edu.hk} \\
  \texttt{sakurapeng@tencent.com, eraserhuang@tencent.com, yuetangdeng@tencent.com} \\
}

\begin{document}
\maketitle

\begin{abstract}
Code-agent RL often receives weak feedback: rollout-time signals are reliable and executable, but capture only necessary or surface conditions for task success rather than the target semantic predicate. Using agentic compile-fix as the setting, we study signal reshaping for standard GRPO under such feedback. Our central claim is that GRPO's within-group comparison is meaningful only after three kinds of signals are reshaped: outcome rewards recover semantic ranking, process signals localize intra-trajectory credit, and rollouts from the same prompt remain execution-comparable. We operationalize these conditions with a minimal signal-reshaping construction that leaves GRPO's group-normalized advantage construction unchanged: compile-and-semantic layered rewards reshape trajectory ranking, step-level process scores outside group reward normalization reshape within-trajectory update strength, and failure-cause-aware rollout governance reshapes within-group comparability. Experiments show a clear end-to-end gain: full signal-reshaped GRPO improves strict compile-and-semantic accuracy from the base model's zero-shot $0.385$ to $0.535$. Controlled comparisons further explain the source of this gain: binary rewards remove the compile-only middle tier and degrade trajectory control; on top of layered rewards, process-score weighting further improves accuracy from $0.48$ to $0.53$ and reduces average evaluation steps from $23.50$ to $17.02$. As a boundary comparison, privileged-prompt token-level distillation mainly optimizes local distributional alignment; in long tool-use trajectories, this signal is diluted by non-critical tokens and cannot replace outcome semantics, process credit, or within-group comparability.
\end{abstract}

\section{Introduction}

Code large language models (LLMs) are moving from one-shot code generation toward interactive software repair in real repositories~\citep{jimenez2024swebench,zan2025multiswebench,yang2024sweagent,wei2025swerl,li2026laboratoryrealworldapplicationsbenchmarking}. A common paradigm trains the model as a code agent that interacts with repositories through ReAct-style tool use~\citep{yao2023react} and updates its policy from execution feedback with group-relative reinforcement learning, such as Group Relative Policy Optimization (GRPO)~\citep{shao2024deepseekmath,luo2025deepswe,yang2025kimidev,golubev2025longhorizonswe}. GRPO relies on relative rankings among same-prompt rollouts; if these rankings no longer mainly reflect repair progress, advantage updates can optimize noise or surface shortcuts. Agentic repair makes this premise nontrivial: trajectories are no longer pure text sequences, but long interactions composed of context reading, code editing, compilation checks, and explicit termination; rollout outcomes are also jointly determined by tool definitions and execution resources. This means that execution-environment jitter, and whether execution feedback is sufficient to characterize true repair success, can affect the optimization target itself.

This within-group comparison problem is especially pronounced in \emph{weak-feedback agentic repair}. Weak feedback denotes online rollout signals that check only necessary or surface conditions, not the target semantic predicate. We use compile-fix as the study setting: an uncompilable patch cannot be correct, but a compiling patch may still delete needed code, add an unsafe stub, or bypass the failing path. During training, no executable tests are available, and literal similarity to a reference patch is not correctness. Thus, neither the test oracle in SWE-bench~\citep{jimenez2024swebench} nor the oracle-patch diff reward in SWE-RL~\citep{wei2025swerl} can serve directly as an online training reward. Compile-fix is therefore a typical weak-feedback task: execution feedback is informative and necessary, but not sufficient, for semantic correctness. Used directly for GRPO's same-prompt ranking, within-group comparison can drift from true repair progress. Signal reshaping aims to realign these comparisons with repair intent.

We view weak-feedback agentic repair first as a \emph{signal reshaping} problem, not a call to rewrite the reinforcement-learning objective. GRPO relies on same-prompt rollout comparison, but this comparison is useful only when reward differences mainly arise from policy behavior. In our setting, this requirement yields three signal conditions: 1) \textbf{outcome-semantics failure}, where stable online signals often capture only surface acceptability and, when used as $\{0,1\}$ rewards, induce syntactic shortcuts (reward hacking); 2) \textbf{diffuse process credit}, where a long-horizon outcome reward cannot separate effective repair steps from uninformative repetition, diluting the semantic signal; and 3) \textbf{broken within-group comparability}, where infrastructure jitter, such as compilation queues or container initialization failures, and policy degeneration, such as no tool calls or catastrophic repetition, can mix with genuine repair failures in the same zero-reward bin and distort rankings.
Thus, signal reshaping in weak-feedback agentic reinforcement learning reduces to two requirements: \emph{(A) rewards must reflect repair intent at the outcome level and support useful process credit; and (B) rollouts entering within-group comparison must be execution-comparable.} We call modifying training signals around the existing GRPO update \emph{signal reshaping}.

We use signal reshaping as a minimal intervention to test these requirements. A layered compile-and-semantic reward assigns $0/0.5/1$ to non-executable, compile-only, and semantically correct repairs, restoring outcome ranking across trajectories. Step-level process scores serve as token-level loss weights that redistribute update strength within each rollout. Failure-cause-aware rollout governance masks or localizes non-learnable exceptions before incomparable samples pollute same-group ranking. The method is not a new GRPO objective; it reshapes the signals existing objective can learn from.

Experiments test signal reshaping layer by layer. End-to-end, full signal-reshaped GRPO improves strict compile-and-semantic accuracy from the base model's zero-shot $0.385$ to $0.535$. Controlled comparisons first test outcome semantics: a pure compilation reward makes the compile rate spike early and then fall back, without improving compile-and-semantic correctness. Under fixed rollout governance, layered rewards preserve roughly $0.30$ mass at the intermediate $R=0.5$ tier, whereas removing this tier jointly degrades trajectory control and termination. From the same post-fork checkpoint, step-level process scores improve accuracy from $0.48$ to $0.53$ and reduce average evaluation steps from $23.50$ to $17.02$, showing that process credit improves repair efficiency rather than merely trading longer exploration for success.

These results show that signal reshaping can recover trainable GRPO comparisons. To test whether token-level dense supervision can replace step-level process credit, we further ask whether training-only privileged prompts can improve an unprompted policy through distillation. Privileged-prompt $\pi$-Distill and on-policy self-distillation (OPSD)~\citep{penaloza2026pi} both underperform the stable GRPO baseline: $\pi$-Distill preserves tool use but drifts toward premature \texttt{finish}, while OPSD exhibits low entropy, enlarged gradients, and tool-level degradation. This negative result suggests that Kullback--Leibler (KL)-style local distribution matching provides dense token-level supervision, but the signal is too diffuse for long code-agent trajectories; see Appendix~\ref{app:distill_failure}.

Figure~\ref{fig:overview} summarizes the framework: a weak-feedback execution environment produces multi-turn rollouts; layered rewards reshape outcome ranking, step-level process scores reshape intra-trajectory credit, and rollout governance reshapes within-group comparability. The $\pi$-Distill and OPSD~\citep{penaloza2026pi} branch on the right provides a boundary comparison for the token-level KL path.

Our main contributions are:
\begin{itemize}[leftmargin=1.5em,itemsep=0pt,topsep=1pt]
    \item We formulate GRPO training in weak-feedback agentic repair as a signal-reshaping problem for within-group comparison, identifying three signal conditions: semantically ordered outcomes, localized process credit, and execution-comparable same-prompt rollouts.
    \item We present a concrete signal-reshaping method that leaves GRPO's group advantage unchanged: layered rewards restore outcome ranking, step-level process scores reallocate within-trajectory update strength, and failure-cause-aware governance keeps same-group rollouts comparable.
    \item We validate these reshaped signals in compile-fix: layered rewards restore stable training, step-level process scores improve accuracy and efficiency, and token-level privileged distillation shows that local distribution matching cannot replace step-level process credit.
\end{itemize}

The rest of the paper formalizes the setting (Section~\ref{sec:prelim}), presents the construction (Section~\ref{sec:method}), and tests the predictions (Section~\ref{sec:experiments}). Related work, limitations, prompts, and the full distillation analysis appear in the appendix.

\begin{figure}[t]
    \centering
    \includegraphics[width=\linewidth]{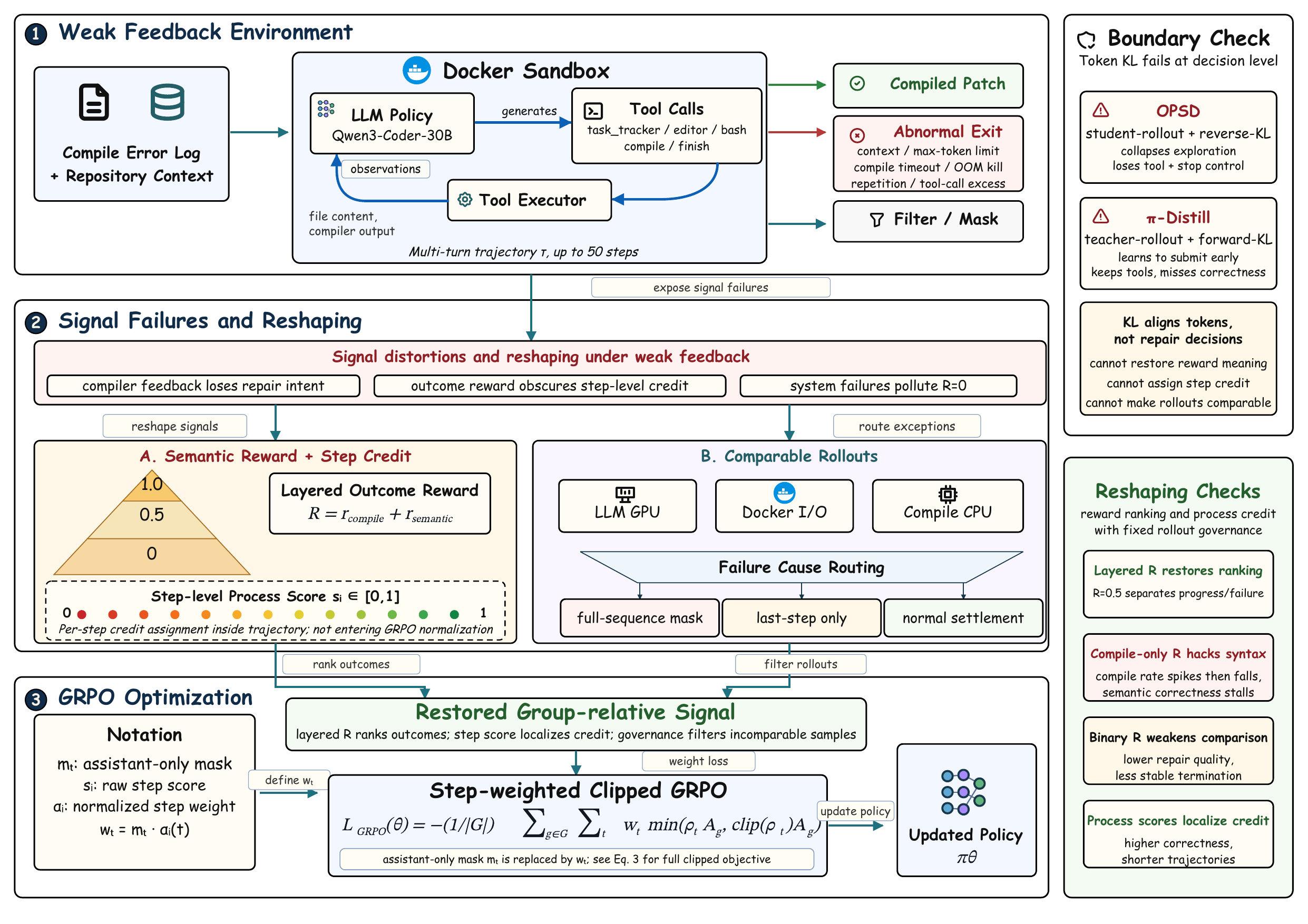}
    \caption{Signal-reshaping framework for GRPO in weak-feedback agentic repair: layered rewards reshape outcome ranking, step-level process scores reshape intra-trajectory credit, and rollout governance reshapes within-group comparability.}
    \label{fig:overview}
\end{figure}

\section{Preliminaries}
\label{sec:prelim}

This section fixes notation for weak feedback, multi-turn trajectories, assistant-only masks, GRPO, and token-level OPD.

\paragraph{Weak feedback.}
Let $C(\tau)$ denote the execution-level signal observable during rollout and $S(\tau)$ denote the target semantic success predicate. Feedback is \emph{weak} when $C$ is reliable and available online, but captures only a necessary condition for $S$, not a sufficient one. In compile-fix, $C$ is compilation success: $C(\tau)=0$ rules out a valid repair, but $C(\tau)=1$ does not guarantee that the patch repairs the intended error. Semantic correctness $S$ must therefore be supplied by a separate judgment. The issue is not observation noise, but signal incompleteness: online execution feedback is informative, yet insufficient to define the task predicate that the policy should optimize.

\subsection{Multi-turn Agentic RL and Assistant-Only Mask}
\label{sec:prelim_agentic}

We model agentic code repair as a finite-horizon Markov decision process (MDP) $\langle \mathcal{S}, \mathcal{A}, D, R, T \rangle$. The state $\mathcal{S}$ contains repository state and conversation history; $\mathcal{A}$ contains natural-language outputs and tool calls; $D$ is the external runtime; $R$ is observed after termination; and $T$ is the maximum number of interaction steps.

Given an initial prompt and repository state $x$, the policy $\pi_\theta$ interacts with the environment to produce a multi-turn trajectory
\begin{equation}
\tau = (a_1, o_1, a_2, o_2, \dots, a_{T'}, o_{T'}),
\label{eq:multiturn-traj}
\end{equation}
where $a_i$ are model actions, $o_i$ are environment observations, and $T'\leq T$. The objective is to maximize $\mathbb{E}_\tau[R(\tau)]$.

\paragraph{Assistant-only mask.} Serialized trajectories contain model generations and environment echoes, but only model tokens are determined by $\pi_\theta$. Let $m_t\in\{0,1\}$ mark whether token $t$ belongs to an assistant generation; $m_t=1$ tokens enter the policy gradient, while $m_t=0$ echoes are masked~\citep{jin2025searchr1,qin2025toolrl}.

\subsection{GRPO}
\label{sec:prelim_grpo}

GRPO~\citep{shao2024deepseekmath} avoids an explicit value function by comparing multiple samples from the same prompt. For each prompt, $K$ trajectories $\{\tau_k\}_{k=1}^K$ are sampled, and scalar rewards $R_k$ are group-normalized into trajectory-level advantages:
\begin{equation}
\hat A_k = \frac{R_k - \bar R}{\mathrm{std}(\{R_i\}_{i=1}^K)}, \qquad
\bar R = \tfrac{1}{K}\textstyle\sum_{i=1}^K R_i,
\label{eq:grpo-adv}
\end{equation}
which are broadcast to all assistant tokens in the corresponding trajectory, denoted by $\hat A_t \equiv \hat A_{k(t)}$. The policy is updated with the clipped policy-gradient objective:
\begin{equation}
\mathcal{L}_{\text{GRPO}}(\theta) =
-\,\mathbb{E}_{x,\,\{\tau_k\}_{k=1}^{K}}
\frac{1}{\max(1,\sum_t m_t)}
\sum_{t} m_t\,\min\!\Big(
\rho_t(y_t)\, \hat A_t,\;\; \mathrm{clip}\big(\rho_t(y_t),\, 1-\epsilon_\text{lo},\, 1+\epsilon_\text{hi}\big)\hat A_t
\Big),
\label{eq:grpo-loss}
\end{equation}
where $y_t$ is response token $t$ and $\rho_t(y_t) = \pi_\theta(y_t\mid \cdot)/\pi_{\theta_{\text{old}}}(y_t\mid \cdot)$. KL and entropy regularization are added in the standard way (full form in Appendix~\ref{app:grpo-hparams}).

\paragraph{Within-group comparison prerequisites.} Equations~\eqref{eq:grpo-adv}--\eqref{eq:grpo-loss} assume that $R$ is a semantically meaningful measure of task success and that the $K$ trajectories sampled from the same prompt are comparable at the execution level. In weak-feedback, long-horizon agentic tasks, these assumptions need not hold; terminal advantages are also broadcast over many assistant tokens, motivating process-level credit assignment.

\subsection{Token-Level On-Policy Distillation with Privileged Prompts}
\label{sec:prelim_opd}

OPD~\citep{agarwal2024onpolicy} matches a teacher distribution on trajectories from the current policy. We use privileged-hint $\pi$-Distill and OPSD~\citep{penaloza2026pi} as boundary comparisons: $\pi$-Distill transfers from hint-conditioned teacher trajectories to a hint-free student, whereas OPSD scores hint-free student trajectories with the hint-conditioned teacher. Both inject masked-sum token KL as a GRPO reward penalty, with directions $D_{\mathrm{KL}}(\pi^T\|\pi^S)$ and $D_{\mathrm{KL}}(\pi^S\|\pi^T)$. This path assumes teacher-preferred behavior aligns with student-feasible behavior and that critical decisions are not diluted by low-information tokens. Full objectives are in Appendix~\ref{app:distill-impl}; the corresponding empirical analysis is in Appendix~\ref{app:distill_failure}.

\section{Method}
\label{sec:method}

This section presents the signal-reshaping construction: layered rewards reshape outcome ranking, step-level scores reshape intra-trajectory credit, and rollout governance reshapes within-group comparability. Table~\ref{tab:design-map} maps the three components to the corresponding signal distortions and reshaping mechanisms.

\begin{table}[t]
    \centering
    \footnotesize
    \setlength{\tabcolsep}{4pt}
    \renewcommand{\arraystretch}{1.15}
    \caption{Signal conditions, signal distortions, and reshaping mechanisms.}
    \label{tab:design-map}
    \begin{tabular}{@{}p{0.20\linewidth} p{0.28\linewidth} p{0.34\linewidth} c@{}}
        \toprule
        \textbf{Condition} & \textbf{Signal distortion} & \textbf{Reshaping mechanism} & \textbf{Section} \\
        \midrule
        \multirow{4}{=}{(A) Semantic\\reward signal}
            & Surface acceptability conflates true repair
            & Layered reward $R=r_{\text{compile}}+r_{\text{semantic}}$; compile-only $\leq0.5$
            & \S\ref{sec:reward} \\
        \addlinespace[3pt]
            & Outcome advantage cannot localize key steps
            & Step score $s_i$ induces token weight $w_t=m_t\alpha_{i(t)}$
            & \S\ref{sec:reward} \\
        \midrule
        \multirow{4}{=}{(B) Comparable\\within-group execution}
            & System bottlenecks reshape sampling
            & Decouple concurrency by resource type
            & \S\ref{sec:rollout} \\
        \addlinespace[3pt]
            & Abnormal exits and genuine failures are mixed into $0$-reward failures
            & Route by failure cause: mask, retain last step, or terminate normally
            & \S\ref{sec:rollout} \\
        \bottomrule
    \end{tabular}
    \vspace{-0.6em}
\end{table}

\subsection{Task Formalization}
\label{sec:task_setup}

Following Section~\ref{sec:prelim_agentic}, each compile-fix sample contains a failure report, repository context, tool interfaces, and a developer repair from which we derive the ground-truth (GT) repair; the GT is not visible during rollout and, in the absence of executable tests, serves as a reference-conditioned semantic proxy for reward construction. The agent runs in a Docker sandbox with file editing (\texttt{str\_replace\_editor}), shell (\texttt{bash}), incremental compilation (\texttt{compile\_code}), task tracking (\texttt{task\_tracker}), and termination (\texttt{finish}) tools; details are in Appendix~\ref{app:env-tools}. At each step, the policy acts until termination or the step limit.

In this setting, the availability of training signals depends on whether a trajectory can be executed to completion. This condition is coupled with system constraints such as compilation queues, container startup, and context budgets; as a result, execution-level comparability within a group is not guaranteed by default.

\subsection{Training Data Construction}
\label{sec:data}

The training corpus comes from compilation-failure events in an internal build system. Each raw record contains a static repository snapshot, build configuration, compilation error log, and developer repair. Directly using the developer repair as a supervision target introduces two sources of noise: (i) human commits often merge the error fix with feature changes, refactoring, or style cleanup, so only a small subset of changes actually eliminates the error; and (ii) the same root cause, such as a cascading upstream interface change, can appear repeatedly in near-duplicate form, letting a few frequent errors dominate the data distribution.

We therefore construct \emph{minimal sufficient repairs} through the following steps: (i)~\textbf{Repair snapshot alignment}: align each failed build with the triggering commit and the subsequent repair commit, extract a normalized unified diff, and obtain a tuple of (compilation error, developer repair, build environment) with repository coordinates; (ii)~\textbf{Structuring and filtering}: deduplicate by error-log fingerprints, classify errors into coarse/fine categories covering syntax, type, declaration, function, linking, and related failures for stratified sampling, and remove pure-deletion repairs that are more often refactoring than error repair; (iii)~\textbf{Minimal sufficient repair refinement}: constrain each developer repair to a strict subset that is both necessary and sufficient, removing only irrelevant changes and introducing no new rewrites, and verify it end-to-end through isolated incremental compilation. The resulting minimal sufficient repair is the GT repair used in later semantic judgment and distillation, keeping unrelated changes from contaminating the semantic judgment reference and providing a clean abstraction source for distillation; (iv)~\textbf{Recent-change clues}: for each reported source file, trace the most recent pre-failure diff, which often identifies root causes such as interface changes, and provide it in both training and evaluation prompts; and (v)~\textbf{Split}: stratify by primary error type and refined repair size to obtain disjoint training and evaluation sets, then further filter the training set to remove overly simple samples, reducing cases where all $K$ rollouts within a group succeed simultaneously and cause advantage collapse. After refinement, each minimal sufficient repair is called the GT repair and serves as the semantic judgment reference rather than a token-level target for imitation.

\subsection{Layered Rewards and Step-Level Process Scores}
\label{sec:reward}

The build system indicates whether code is accepted, not whether the original error is correctly repaired. Used directly as a binary GRPO reward, it can rank shortcuts such as commenting out error sites, stubs, or unconditional \texttt{return} statements together with genuine fixes~\citep{amodei2016concrete,skalse2022rewardhacking}. The reward must therefore separate three cases that binary compilation feedback conflates: no executable repair, executable but semantically wrong repair, and executable semantically correct repair.

\paragraph{Scalar reward: layered $\{0, 0.5, 1\}$.}
\begin{equation}
R(\tau) = r_{\text{compile}}(\tau) + r_{\text{semantic}}(\tau),
\end{equation}
where $r_{\text{compile}} \in \{0, 0.5\}$ indicates whether incremental compilation succeeds, and $r_{\text{semantic}} \in \{0, 0.5\}$ is assigned by a strong model (Kimi-K2.5~\citep{team2026kimik25}) that compares the semantic consistency between the agent patch and the GT repair patch. The criterion is functional equivalence to the GT repair, not lexical similarity. Semantic checking is performed only after compilation succeeds: a compilation failure receives $R=0$, a compilation success with semantic inconsistency receives $R=0.5$, and a compilation success with semantic consistency receives $R=1$.

Separating compilation success from semantic alignment and assigning $0.5$ to each has two effects. First, compile-only shortcuts receive at most partial credit, so full reward is reserved for repairs consistent with the GT repair. Second, the intermediate $0.5$ level preserves within-group ranking signal early in training, when fully semantic repairs are still sparse.

\paragraph{Step-level process scores.}
The scalar reward $R$ is sparse: critical edits and repetitive operations inherit the same trajectory advantage. Motivated by process supervision for long reasoning and program generation~\citep{lightman2024prm,wang2024mathshepherd,dou2024stepcoder}, we introduce a step-level score $s_i \in [0,1]$ for directional correctness and information gain.

The design follows three constraints: (i)~$s_i$ only redistributes gradient \emph{magnitude} and does not enter GRPO scalar-reward normalization; (ii)~scores are assigned only to trajectories that compile successfully, avoiding attribution on non-executable code; and (iii)~the weights are applied on top of the assistant-only mask $m_t$. In the simplified view, each step score is converted into a loss weight relative to the trajectory average, then written to that step's assistant-token mask. Concretely, let $n_i$ be the number of active assistant tokens in step $i$ and
\begin{equation}
\bar{s}=\frac{\sum_i n_i s_i}{\sum_i n_i},\qquad
\alpha_i^+=\frac{s_i}{\bar{s}},\qquad
w_t=m_t\,\alpha_{i(t)} .
\label{eq:step-weight}
\end{equation}
Positive-advantage trajectories use $\alpha_i=\alpha_i^+$. Negative-advantage trajectories use $\alpha_i=c\,\max(2-\alpha_i^+,0.1)$, with $c$ normalizing the weights so that $\sum_i n_i\alpha_i=\sum_i n_i$. The sign is reversed to invert step weights when penalizing bad trajectories: low-scoring steps receive stronger negative updates, while high-scoring steps are penalized less. Thus active loss-mask mass is preserved and gradients are only redistributed across steps: $R$ ranks trajectories, while $s_i$ allocates credit within a trajectory.

\paragraph{Scoring criteria and information gain.}
Process scores are emitted step by step by a strong model conditioned on the full trajectory context. The judge follows the same semantic-equivalence, functional-consistency, and behavior-preservation criteria as outcome judging, but lowers the comparison granularity from the full patch to a single tool call. To suppress scoring drift, the judge focuses only on modifications related to the target compilation error and explicitly ignores unrelated changes in the GT. It also follows an information-gain principle: low-gain steps such as repeatedly viewing already inspected files are penalized, while steps that trigger compilation verification receive a higher base score to encourage the policy to seek executable feedback.

\subsection{GRPO Optimization}

The two-level supervision changes only the token coefficient in standard GRPO: Eq.~\eqref{eq:grpo-loss} replaces $m_t$ with the effective coefficient $w_t=m_t\alpha_{i(t)}$ from Eq.~\eqref{eq:step-weight}. Disabling process scores sets $\alpha_{i(t)} \equiv 1$, recovering the assistant-only mask.

The configuration follows two principles. First, each prompt samples $K=8$ trajectories with rollout temperature $1$ to keep within-group candidates diverse. Second, asymmetric clipping, with a looser upper than lower bound, follows large-scale GRPO/RL stabilization practice~\citep{guo2025deepseekr1,yu2025dapo}: it permits larger updates toward beneficial exploration while making degenerate updates more conservative. A small $\lambda_{\text{KL}}$ anchors the policy to the frozen reference model; details are in Appendix~\ref{app:grpo-hparams}.

GRPO itself is not the algorithmic contribution. The contribution is signal placement and rollout governance: GT repairs provide semantic references, layered $R$ gives outcome ranking, $s_i$ gives process credit without changing group normalization, and governance keeps within-group samples comparable.

\subsection{Rollout Governance}
\label{sec:rollout}

Effective GRPO learning also requires that the $K$ rollouts in a group have the same kind of execution opportunity. Prior work on multi-turn agentic RL shows that environment echoes, tool calls, and rollout governance directly affect the stability of learnable signals~\citep{jin2025searchr1,qin2025toolrl,wang2025ragen}; in compile-fix, this issue is amplified by container setup and compilation queues.

\paragraph{Decoupling sampling concurrency by execution-resource dimension.} Rollouts consume three heterogeneous resources: policy inference on GPUs, sandbox operations under host scheduling, and incremental compilation under CPU and memory pressure. A single concurrency controller lets compilation dominate latency, while excessive compilation concurrency causes contention and timeouts. We therefore decouple the three resource types and tighten the compilation limit, preventing compilation from becoming the rollout bottleneck while improving resource utilization.

\paragraph{Routing trajectories by failure cause.} Without failure-cause routing, $0$-reward samples mix repair failures, policy degeneration, and trajectories without valid execution opportunity. We separate \emph{unlearnable exceptions} from \emph{informative failures} by exit source and use three handling strategies:
\begin{itemize}[leftmargin=1.5em,itemsep=0pt,topsep=1pt]
    \item \textbf{Full-trajectory masking}: exceptions that have no causal relation to model decisions and therefore should not enter the training objective. These include environment initialization failures; context overflow during generation, inference interruption, truncation caused by reaching the per-step output limit, and early termination triggered by repeated compilation timeouts; as well as final compilation exceptions caused by the execution environment rather than repair semantics, including compilation timeouts and out-of-memory cases where the compilation process is killed by the system. The entire response is removed from the training objective.
    \item \textbf{Last-step retention}: patterns that arise from sampling itself and whose degeneration signal is concentrated at the end of the trajectory. Currently these include out-of-bound concurrent tool calls within a single turn and catastrophic repetition, both of which are detected during generation and trigger early termination. For such samples, the preceding trajectory is masked and only the final step is retained as a penalty sample, aligning negative feedback with the position that actually reflects policy degradation.
    \item \textbf{Normal termination}: all remaining valid trajectories proceed to compilation and semantic judgment.
\end{itemize}

\section{Experiments}
\label{sec:experiments}

This section evaluates the signal reshaping from Sections~\ref{sec:prelim}--\ref{sec:method}. As the end-to-end result, the full two-stage GRPO trajectory improves strict compile-and-semantic accuracy from $0.385$ at the first evaluation point (\texttt{rollout/step}$=0$, i.e., the base model zero-shot) to $0.535$, an absolute gain of $+15.0$pp. Q1 tests whether layered rewards provide learnable trajectory-level ordering; Q2 then tests whether step-level process scores improve intra-trajectory credit assignment.

\subsection{Research Questions and Compared Systems}

We do not ablate rollout-governance components individually. Preliminary runs showed that disabling any single component often breaks execution comparability early in training, causing training collapse or unstable convergence and making reward and credit-assignment effects hard to interpret cleanly. We therefore keep the rollout pipeline fixed and isolate two training-signal conditions:
\begin{enumerate}[leftmargin=1.5em,itemsep=0pt,topsep=1pt]
    \item \textbf{Q1 (Reward semantics).} Are compile-and-semantic layered rewards the reward-semantic condition needed for stable GRPO training?
    \item \textbf{Q2 (Process credit).} Once layered rewards produce learnable trajectories, can step-level process scores further improve GRPO updates through finer-grained intra-trajectory credit assignment?
\end{enumerate}

The four GRPO configurations share the same base model, rollout system, data split, and training hyperparameters, and differ only in reward form and whether step-level process scores are used:
\begin{itemize}[leftmargin=1.5em,itemsep=0pt,topsep=1pt]
    \item \textbf{GRPO-compile-only}: uses a further weakened binary reward ($0/1$, $R=1$ iff compilation succeeds), removing semantic checks from the reward.
    \item \textbf{GRPO-early}: layered rewards ($0/0.5/1$) with the standard assistant-only loss mask.
    \item \textbf{GRPO-full (main method)}: continued from the GRPO-early fork point with the same layered reward plus step-level process-score weighting.
    \item \textbf{GRPO-binary}: continued from the same fork point with binary rewards ($0/1$, $R=1$ iff compilation succeeds and the repair is semantically correct), collapsing compile-only partial success into failure.
\end{itemize}

\paragraph{Protocol summary.} All methods use the Docker-based compile-fix environment in \S\ref{sec:task_setup}, Qwen3-Coder-30B-A3B-Instruct as the base model, a $1110$-sample training set, and a fixed $400$-sample hint-free evaluation split; full hyperparameters are in Appendix~\ref{app:grpo-hparams}.

\begin{figure}[t]
    \centering
    \includegraphics[width=\linewidth]{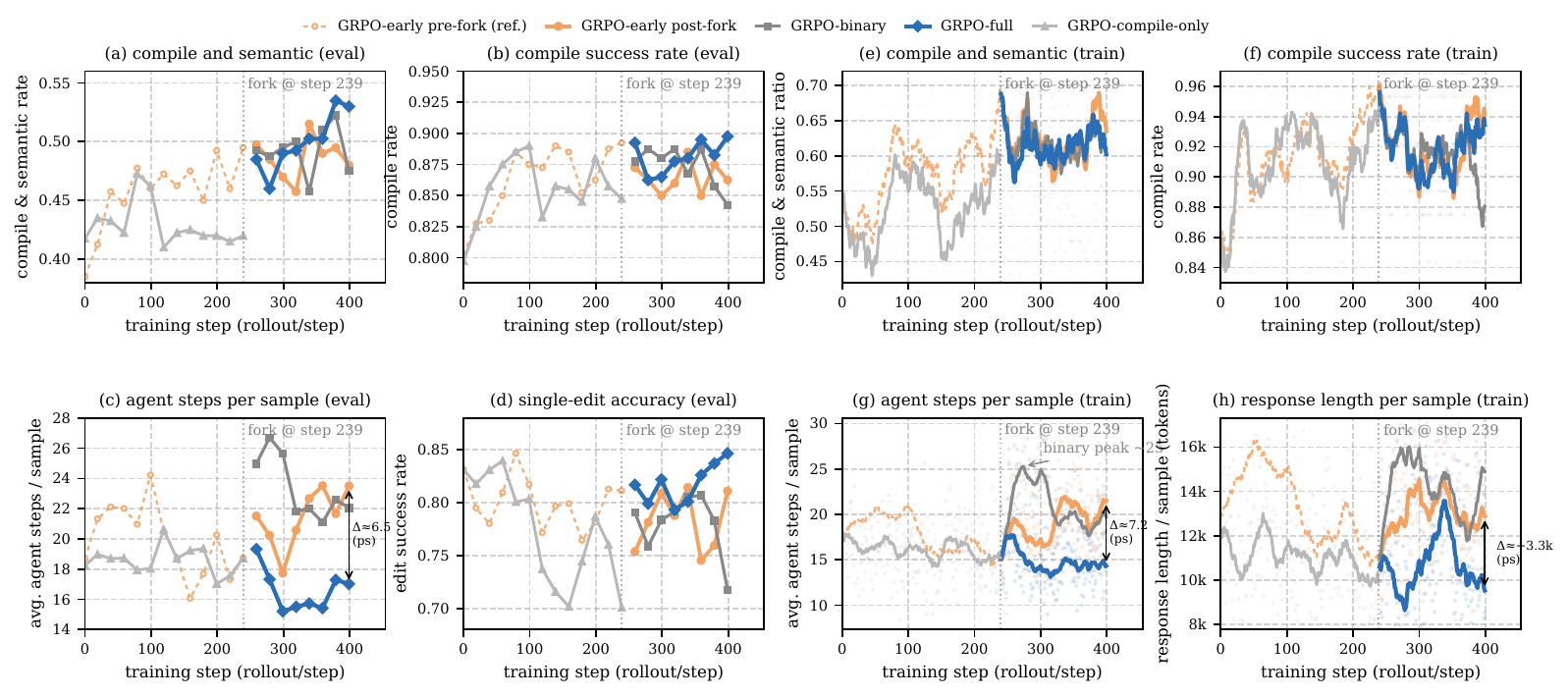}
    \caption{Four GRPO settings on a unified $\texttt{rollout/step}$ axis. GRPO-early/binary/full share the fork at \texttt{rollout/step}$=239$; GRPO-compile-only and GRPO-early (pre-fork segment) both start independently from the base model. Panels (a)--(d): fixed $400$-sample evaluation; (e)--(h): train-time signals; curves use a $20$-step centered rolling average.}
    \label{fig:grpo_combined}
    \vspace{-0.8em}
\end{figure}

\subsection{Q1: Layered Rewards Support Stable Training}
\label{sec:grpo_comparison}

\begin{figure}[t]
    \centering
    \begin{subfigure}[t]{0.49\linewidth}
        \centering
        \includegraphics[width=\linewidth]{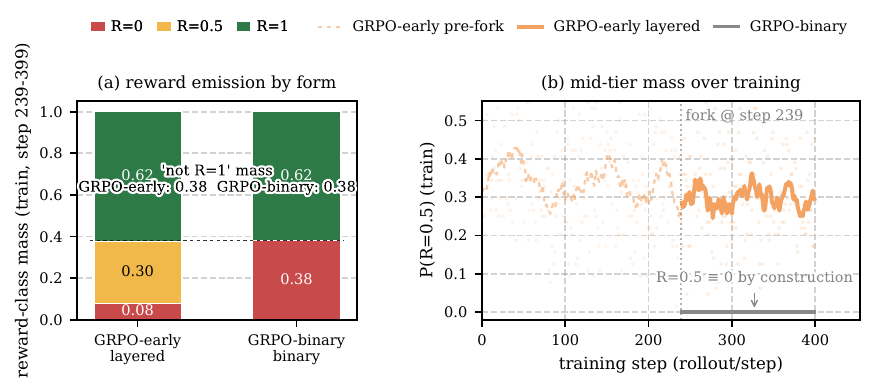}
        \caption{Reward emission for layered vs.\ binary rewards. Left: reward-level distribution in the shared window; right: middle-tier training dynamics.}
        \label{fig:reward_distribution}
    \end{subfigure}
    \hfill
    \begin{subfigure}[t]{0.49\linewidth}
        \centering
        \includegraphics[width=\linewidth]{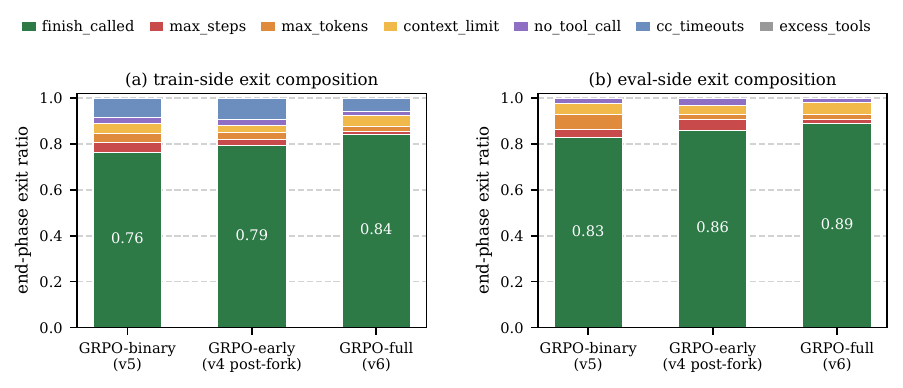}
        \caption{Exit composition at the tail of the shared window. Left: training rollouts; right: $400$-sample evaluation split.}
        \label{fig:exit_reason_grpo}
    \end{subfigure}
    \caption{Mechanism evidence for outcome-level reward semantics: (a) binary rewards remove the $R=0.5$ tier; (b) weaker comparison semantics coincide with worse termination.}
    \label{fig:reward_and_exit_grpo}
    \vspace{-1.0em}
\end{figure}

\begin{takeaway}{E1 (Reward semantics): binary rewards remove the middle signal needed for within-group comparison.}
In the shared post-fork window, layered GRPO assigns about $0.30$ reward mass to $R=0.5$, corresponding to executable but semantically incomplete repairs; binary GRPO lacks this middle tier and folds these samples into $R=0$. Figure~\ref{fig:reward_and_exit_grpo}(a) shows that the middle tier preserves the ordering information between compile-level progress and complete failure.
\end{takeaway}

\begin{takeaway}{E2 (Binary-reward degradation): repair quality decreases with less stable termination behavior.}
At the end of the post-fork comparison, GRPO-binary reaches $0.475$ compile-and-semantic accuracy, below GRPO-early at $0.480$. Its auxiliary metrics decline in the same window: compilation success drops from $0.8775$ to $0.8425$, and single-edit success\footnotemark drops from $0.791$ to $0.717$. Figure~\ref{fig:grpo_combined}(a)--(d) and (g)--(h) show that lower quality is accompanied by longer trajectories; Figure~\ref{fig:reward_and_exit_grpo}(b) gives the same diagnosis: GRPO-binary has the lowest normal-termination ratio and a higher abnormal-exit burden than GRPO-full.
\end{takeaway}
\footnotetext{Single-edit success is the fraction of \texttt{str\_replace\_editor} calls that apply an exact string replacement and return success; it measures tool-level edit execution quality.}

\begin{takeaway}{E3 (Execution-signal reward hacking): pure compile-only binary rewards push the policy toward syntactic shortcuts.}
GRPO-compile-only and GRPO-early share the pre-fork window; only the reward differs, determined solely by \textit{is\_compiled}. The optimized compile rate spikes early and then declines, with less stability than GRPO-early (Figure~\ref{fig:grpo_combined}(f)); meanwhile, compile-and-semantic correctness stagnates or slightly decreases on training and evaluation signals (Figure~\ref{fig:grpo_combined}(a,e)), and tool-level single-edit success\footnotemark[\value{footnote}] also drops (Figure~\ref{fig:grpo_combined}(d)). This matches the outcome-semantics failure from the introduction: execution feedback is necessary but insufficient for semantic correctness. Used directly as a $\{0,1\}$ reward, it lets within-group comparison favor syntactic shortcuts that bypass semantic checks, leaving compile success without stable support. Adding semantic-level outcome signals strengthens the target, steadily improves compile-and-semantic success, and stabilizes compile success. Outcome rewards therefore need semantic-level reshaping.
\end{takeaway}

\begin{takeaway}{E4 (Stable convergence): layered rewards and rollout governance provide a trainable comparison.}
Layered rewards provide outcome semantics, and fixed rollout governance keeps within-group samples comparable; together, they support GRPO's within-group comparison. The GRPO-early training curves in Figure~\ref{fig:grpo_combined}(e)--(f) enter a stable range after early variation, without sustained oscillation, divergence, or collapse toward low-success rollouts.
\end{takeaway}

Thus, Q1 asks whether the task can be learned stably; Q2 tests whether step-level process scores further improve intra-trajectory credit assignment under this condition.

\subsection{Q2: Process Scores Improve Correctness and Efficiency}
\label{sec:main_grpo_result}

This section isolates the second-stage mechanism. The GRPO-early post-fork segment and GRPO-full start from the same fork point and use the same layered scalar reward; the only additional signal in GRPO-full is the step-level process score applied to the loss mask.

\begin{takeaway}{E5 (Process-level credit assignment): step weighting improves correctness while shortening trajectories.}
In the post-fork comparison, GRPO-full improves compile-and-semantic accuracy ($0.480\!\to\!0.530$, $+5.0$pp), compilation success ($0.8625\!\to\!0.8975$), and single-edit success\footnotemark[\value{footnote}] ($0.811\!\to\!0.847$) over GRPO-early. These metrics cover tool execution, compilation, and semantic correctness; their joint gain acts on the execution chain rather than only fitting the top-level reward. Average evaluation steps fall from $23.50$ to $17.02$; on training trajectories, late smoothed agent steps fall from $21.17$ to $13.92$, and response length from about $12.9$k to $9.7$k tokens (Figure~\ref{fig:grpo_combined}(g)--(h)). Thus, correctness improves alongside more effective credit assignment over tool steps, not through longer exploration.
\end{takeaway}

\begin{takeaway}{E6 (Termination behavior): process scores shift the policy toward more stable normal termination.}
In the train-side tail, GRPO-full has the highest $\texttt{finish\_called}$ ratio ($0.842$) and lowest $\texttt{max\_steps}$ ratio ($0.014$), ahead of GRPO-early ($0.793/0.027$) and GRPO-binary ($0.764/0.044$). The evaluation tail shows the same ordering, with GRPO-full at $\texttt{finish\_called}=0.888$ and $\texttt{max\_steps}=0.018$. The efficiency gain therefore reflects fewer steps and more normal termination within budget.
\end{takeaway}

Together, Q1 and Q2 address two levels of signal reshaping: layered rewards first provide trajectory-level preferences, and step-level process scores then refine intra-trajectory credit. The main benefit does not come from changing GRPO's group-normalized scalar-reward update form, but from first providing comparison semantics and then improving token-level loss weighting over tool steps.

\section{Conclusion}

This paper treats weak-feedback agentic code repair as a \emph{signal reshaping} problem: the online signal is reliable but semantically incomplete, so GRPO remains effective only when within-group comparisons remain meaningful. Our approach reshapes the training signals around an unchanged objective: outcome ranking, process credit, and within-group comparability.
The evidence supports this ordering as follows. Pure compile rewards make compile rate spike early and then fall back less stably, without turning into compile-and-semantic correctness; layered rewards restore outcome ranking, steadily improve the compile-and-semantic success rate, and provide a more stable semantic support for compile success. Step-level weighting then improves correctness and efficiency. Token-level privileged supervision fails to provide the same signal-level correction, consistent with a mismatch between token-level alignment and tool-level repair decisions. For other weak-feedback agentic tasks, our findings suggest a sharper design principle: when the objective is right but the feedback is incomplete, reshaping the signal around that objective is more effective than adding mismatched supervision.

% \begin{ack}
% Acknowledgments go here.
% \end{ack}

\bibliographystyle{plainnat}
\bibliography{main}

\appendix
\startappendixonlytoc
% -- Appendix Title + Local Table of Contents ---------------------------------
% Read the appendix-only .apt file written after \appendix.
\begin{center}
    {\LARGE\bfseries Appendix for ``Signal Reshaping for GRPO in Weak-Feedback Agentic Code Repair''\par}
\end{center}
\vspace{1.2em}

\begingroup
\setcounter{tocdepth}{2}
\section*{Table of Contents}
\vspace{-0.4em}\noindent\hrulefill\par\vspace{0.6em}
\appendixonlytableofcontents
\vspace{0.4em}\noindent\hrulefill\par
\endgroup
\vspace{1.2em}

\section{Related Work}
\label{sec:related}

Existing automatic program repair work often formulates repair as one-shot patch generation from buggy code to fixed code, including neural patch generation and recent LLM-based repair at function and repository scales~\citep{chen2021sequencer,jiang2021cure,silva2025repairllama,pan2026reporepair,li2026vulkey}. The central object remains a single generated patch, and training and evaluation mainly ask whether that patch is correct, rather than modeling a repair process with tool interaction, multi-step decision-making, and trajectory-level credit assignment.

Repository-level benchmarks and tool interfaces first change the object of repair: from generating a patch to completing an executable repair trajectory in a real repository. SWE-bench asks models to modify repository code for real GitHub issues~\citep{jimenez2024swebench}, while SWE-agent and OpenHands further emphasize agent-computer interfaces, file editing, shell/test calls, and termination decisions~\citep{yang2024sweagent,wang2024openhands}; the ReAct reasoning-action framework also provides a general paradigm for such multi-turn tool interaction~\citep{yao2023react}. In this setting, the model output is no longer only the final patch, but also includes context reading, error localization, file editing, shell/test invocation, feedback interpretation, and termination. In other words, agentic code repair moves code repair from static patch prediction to a long-horizon decision problem jointly defined by tool interaction, execution feedback, and termination decisions.

The next step is to place such long trajectories inside an RL objective. Existing code RL work usually converts unit tests, execution results, or compilation feedback into scalar rewards and then uses deep reinforcement learning to optimize code generation models~\citep{le2022coderl,liu2023rltf,shojaee2023execution,dou2024stepcoder}; general LLM post-training also includes variants such as PPO-style RLHF~\citep{schulman2017ppo}, DPO~\citep{rafailov2023dpo}, KL distillation~\citep{penaloza2026pi}, and DAPO~\citep{yu2025dapo}. In recent long-chain reasoning and code-agentic RL, GRPO/R1-style same-prompt multi-sample relative optimization has become a common backbone~\citep{shao2024deepseekmath,guo2025deepseekr1,wei2025swerl,luo2025deepswe,golubev2025longhorizonswe}; we therefore take GRPO as the main object of analysis. Compared with one-shot patch generation, long trajectories do not by themselves require the three kinds of signal reshaping. These conditions arise from GRPO's within-group comparison mechanism: rollouts for the same prompt yield learnable advantages only when the outcome reward provides reliable semantic ordering, the process signal can localize effective actions within the trajectory, and rollout execution does not lose comparability because of resource contention or system exceptions. We organize related work accordingly: outcome rewards define ``which trajectory is better,'' process weights define ``which tokens are more important within the same trajectory,'' and rollout governance preserves sample comparability before group normalization.

\paragraph{Outcome-level reward semantics.}
CodeRL, RLTF, execution-based code RL, and StepCoder mainly convert tests, execution results, or compilation outcomes into pass/fail rewards~\citep{le2022coderl,liu2023rltf,shojaee2023execution,dou2024stepcoder}. When the test oracle is strong enough, pass/fail can approximate semantic correctness; in compile-fix, however, successful compilation is only a necessary condition. A compile-only binary reward collapses syntactic executability and functional semantic repair into the same positive class, and therefore cannot distinguish patches that merely remove compilation errors from patches that actually repair the intended behavior. Another line of work uses reference-patch similarity or repository-level benchmark outcomes as proxy signals~\citep{jimenez2024swebench,zan2025multiswebench,wei2025swerl,li2026laboratoryrealworldapplicationsbenchmarking}, but patch similarity can conflate implementation style with functional equivalence.

Our layered outcome reward changes only the result semantics used for within-group comparison; it does not change the form of GRPO advantages. Compared with compile-only pass/fail, we keep an intermediate tier: non-compiling repairs receive $0$, compiling but semantically inconsistent repairs receive $0.5$, and repairs semantically consistent with the minimal sufficient ground-truth repair receive $1$. Thus, GRPO's within-group comparison no longer observes only whether compilation passes, but can also use the partial progress of a repair that compiles while remaining semantically incomplete. Compared with patch-similarity rewards, $r_{\text{semantic}}$ focuses on reference-conditioned functional consistency rather than surface edit similarity. The layered reward therefore restores semantic-correctness ordering for outcome rewards in weak-feedback compile-fix.

\paragraph{Process reward models.}
PRMs and their automated variants mitigate credit assignment in multi-step decision making by scoring intermediate reasoning or action steps~\citep{lightman2024prm,wang2024mathshepherd,luo2024omegaprm,zheng2025processbench,yuan2025implicitprm,zhao2026genprm}. Their standard interface is to add step scores as process rewards to the optimization objective. For GRPO, these process rewards are aggregated into the rollout-level scalar reward (or equivalently into the trajectory return), then normalized together with outcome rewards by the same-prompt group mean and variance, thereby changing the advantages of different rollouts~\citep{shao2024deepseekmath,mroueh2025grpo}. As a result, reward scales and variances from different sources can interact. Recent multi-reward GRPO analyses explicitly show that vanilla GRPO advantages can be dominated by high-variance reward components, potentially ignoring lower-variance objectives and inducing reward hacking~\citep{ichihara2026mogrpo}; summing multiple reward sources before group normalization can also collapse distinctions among reward combinations, reduce the resolution of the training signal, and lead to worse convergence or early failures~\citep{liu2026gdpo}. In addition, when PRM scores enter RFT as sum-form process rewards, models have been observed to hack high-scoring steps and even collapse early in training~\citep{cheng2025stopsummation}. Thus, in long-horizon execution tasks, directly folding process scores into the scalar reward can amplify process-scoring noise into group-relative preferences and destabilize training.

We separate outcome ranking from process attribution. The layered compile-and-semantic reward $R$ ranks different rollouts under the same prompt; the step-level process score $s_i$ does not enter GRPO's within-group scalar reward and does not change cross-trajectory advantages. It is converted into token loss weights only inside compilation-successful trajectories with valid process scores, increasing update strength on key edit and verification steps while down-weighting repeated inspection, uninformative explanation, and other low-information operations. Thus, our claim is not that the process judge contains more semantic information than a PRM, but that it uses a different optimization interface: $s_i$ is not an additional reward source, and only reweights token gradients after $R$ has fixed the update direction. This preserves the density of semantic process supervision while avoiding the redefinition of trajectory-level ranking by step count, tool-call frequency, or score scale through group normalization~\citep{amodei2016concrete,skalse2022defining}.

\paragraph{Comparability of long-trajectory rollouts.}
GRPO~\citep{shao2024deepseekmath} and DAPO~\citep{yu2025dapo} are mainly validated in reasoning settings with no external execution environment or weak environment coupling~\citep{guo2025deepseekr1}, and recent analyses show that group-relative comparison itself can introduce bias~\citep{liu2025drgrpo}. In agentic tasks, prior work improves rollout signals through within-group variance filtering~\citep{wang2026ragen2}, environment-echo token masking~\citep{jin2025searchr1}, or tool-call-level reward organization~\citep{qin2025toolrl,wei2025swerl}. Our comparability concern is closer to execution opportunity: container startup, host scheduling, concurrent compilation, context growth, and system exceptions can make the $K$ rollouts for the same prompt face different execution constraints. Our rollout governance is therefore not a new reward source, but execution control before within-group comparison: concurrency decoupling reduces the effect of system bottlenecks on the sampling distribution, and failure-cause routing separates environment exceptions, sampling degeneration, and ordinary repair failures, thereby preventing environmental noise from contaminating GRPO's within-group ranking.

\paragraph{Granularity of additional supervision.}
On-policy distillation and privileged-prompt supervision have recently been used for mathematical reasoning and short-horizon tool interaction~\citep{agarwal2024onpolicy,xu2025kdrl,song2026opdsurvey,yang2026nemotroncascade2}. These methods implicitly require the teacher-preferred token distribution to guide the student toward better task decisions. In weak-feedback, long-trajectory code-agent settings, this correspondence is unstable: repair success is often determined by a few edit, verification, and termination actions, while many tokens are formatting, explanations, or tool echoes. When $\pi$-Distill and OPSD~\citep{penaloza2026pi} write token-level KL into the trajectory reward through a \texttt{masked-sum}, the KL is diluted by low-information tokens and also changes GRPO's within-group reward scale as trajectory length varies, interfering with the trajectory ranking that should be determined by outcome reward. Our boundary comparison (Appendix~\ref{app:distill_failure}) shows two manifestations of this granularity mismatch: reverse KL tends toward low entropy and degraded termination behavior, while forward KL slides into a premature-finish shortcut; both underperform the GRPO baseline under strict no-hint evaluation~\citep{fu2026revisiting}. Privileged supervision therefore does not replace the process credit provided by step-level process scores; at the same time, it does not itself provide outcome semantics or within-group comparability. Tool-call-level or trajectory-level supervision may be a more suitable direction~\citep{lyu2025score,stein2026gates}.

\section{Limitations}
\label{app:limitations}

This study uses C/C++ compile-fix tasks as the main setting because they combine reliable executable feedback, long tool-use trajectories, and semantically incomplete rewards in a controlled way. The goal is not to cover all code-repair benchmarks, but to isolate when GRPO's within-group comparisons remain trainable under weak feedback. Extending the same signal-reshaping principles to other programming languages, repositories with executable tests, and agentic tasks whose feedback is less tied to compilation is an important direction for further validation.

In the absence of executable test oracles, the reward and evaluation protocol use minimal sufficient repairs and an LLM-based semantic judge to provide scalable semantic feedback. The refinement pipeline only removes unrelated changes from developer patches, does not add or rewrite repairs, and verifies the refined patch by compilation; the semantic judge then checks functional consistency against this verified repair reference. This protocol makes it possible to train and evaluate on realistic compile-fix samples that lack tests. Future work can further calibrate the semantic judgments on public test-based benchmarks or human-audited subsets.

The experiments use one base model and one repair environment family to control for model capability and environment differences in the mechanism analysis. Consistent with Section~\ref{sec:experiments}, the main experiments keep rollout governance fixed as an execution-control condition rather than disabling its components one by one. Preliminary runs indicated that disabling any individual component often breaks execution comparability early in training, leading to training collapse and failure to converge stably. Therefore, the main text compares layered rewards and step-level process scores under a fixed rollout pipeline. Broader model sweeps, public benchmark replications, and component-level governance diagnostics can further characterize the scope of the framework.

For reproducibility, the paper reports the tool interface, prompt templates, reward definitions, rollout handling rules, hyperparameters, aggregate metrics, and diagnostic plots. We plan to provide anonymized or synthetic examples, the semantic-judgment and process-scoring prompts, plotting scripts, and the full configuration needed to reproduce the analysis on compatible compile-fix data.

\paragraph{Compute resources.}
All reported GRPO variants use the same FSDP training stack, Qwen3-Coder-30B-A3B-Instruct base model, $K=8$ rollouts per prompt, global batch size $64$, and fixed evaluation split; the full configuration is given in Appendix~\ref{app:grpo-hparams}. The main costs come from long-context policy inference, Docker sandbox execution, and repeated incremental compilation. Rollout governance therefore separates GPU inference, sandbox I/O, and compilation queues, as detailed in Appendix~\ref{app:rollout-impl}. The wall-clock spans of the four GRPO comparison runs are: GRPO-compile-only $89.2$h, GRPO-binary $90.6$h, GRPO-early $163.9$h, and GRPO-full $115.6$h.

\paragraph{Broader impacts.}
More reliable weak-feedback code-repair agents can reduce developer effort on routine build failures and improve the maintainability of large software systems. At the same time, code-repair agents without semantic validation or signal governance may still produce behavior-changing repairs if they rely only on shallow execution signals. The signal-reshaping approach in this paper is designed to mitigate this shallow-signal optimization problem and to support more controlled engineering deployment. Practical deployment should be staged and should retain sandboxing, provenance tracking, semantic validation, and necessary human code review.

\section{Environment and Tool-Interface Implementation Details}
\label{app:env-tools}

The agent execution environment is based on a containerized repair image. The original repository is mounted read-only inside the container at \texttt{/testbed}, while a writable working copy is provided at \texttt{/workspace/repair}. The tool interface contains five operators: \texttt{str\_replace\_editor} for file reading, inspection, and string replacement; \texttt{bash} for command execution; \texttt{compile\_code} for incremental compilation and compiler-log feedback; \texttt{finish} for explicit termination and repair submission; and \texttt{task\_tracker} for auxiliary task tracking, which does not affect trajectory termination. Before calling \texttt{finish}, the agent must have made at least one successful call to \texttt{compile\_code}. Table~\ref{tab:tools} summarizes the subcommands and return conventions.

\begin{table}[t]
    \centering
    \scriptsize
    \setlength{\tabcolsep}{4pt}
    \renewcommand{\arraystretch}{1.05}
    \caption{Tool interfaces in the Docker-sandboxed repair environment.}
    \label{tab:tools}
    \begin{tabular}{@{}p{0.17\linewidth} p{0.20\linewidth} p{0.55\linewidth}@{}}
        \toprule
        \textbf{Category} & \textbf{Tool} & \textbf{Function and return} \\
        \midrule
        File inspection and editing & \texttt{str\_replace\_editor} & Four subcommands under a single tool: \texttt{view} reads a file by line range or lists a directory to depth two; \texttt{create} writes a new file; \texttt{str\_replace} replaces a uniquely matched string within an optional line range; and \texttt{insert} inserts content after a specified line. Long outputs are truncated according to the budget. \\
        \addlinespace[2pt]
        Command execution & \texttt{bash} & Executes arbitrary shell commands in a persistent shell session inside the container, including navigation, directory inspection, scripted search, and background jobs. It supports soft timeouts and standard-input (STDIN) injection, and returns standard output and exit code, with hard truncation for overlong output. \\
        \addlinespace[2pt]
        Incremental compilation & \texttt{compile\_code} & Triggers an incremental build through \texttt{build.sh} under the current repository state, classifies failure causes into \texttt{code\_error}, \texttt{timeout}, \texttt{out\_of\_memory}, \texttt{system\_limit}, and \texttt{environment\_error}, and returns structured error lists and logs. This is the only executable-feedback source for the outcome-level reward component $r_{\text{compile}}$. \\
        \addlinespace[2pt]
        Task tracking & \texttt{task\_tracker} & Maintains a structured task list through \texttt{view} and \texttt{plan}, with states \texttt{todo}, \texttt{in\_progress}, and \texttt{done}, allowing the policy to explicitly organize repair plans and progress over long multi-turn trajectories. \\
        \addlinespace[2pt]
        Explicit termination & \texttt{finish} & A trajectory-termination action actively emitted by the policy, indicating that the repair is complete and should be settled. After it is triggered, no subsequent actions are accepted and reward computation begins directly. \\
        \bottomrule
    \end{tabular}
\end{table}

The semantic judge uses Kimi-K2.5 as the backend. Its input is \{compilation error description, model-generated patch list, minimal-sufficient GT repair patch list\}, where the GT repair is the minimal sufficient repair defined in Section~\ref{sec:data}. Its output includes intent consistency, confidence, and detailed analysis. Scoring is based strictly on semantic equivalence and functional consistency, without requiring identical code style or implementation, and lexical similarity is not used as the reward. The implementation keeps only two semantic outcomes, \texttt{consistent} and \texttt{inconsistent}: the former maps to $R=1$ under the layered reward, while the latter, skipped semantic checks, or semantic-judge failures retain compilation-only partial credit $R=0.5$ on compiled samples. Semantic checking is retried at most two times during training and at most five times during evaluation, both with exponential backoff. The reliability of this judge is audited on a $400$-sample human-annotated subset stratified by method and judge label in Appendix~\ref{app:judge-validation}.

\paragraph{Minimal sufficient repairs and recent-change clues.}
The data-refinement pipeline only removes content from the original developer patch; it never adds, rewrites, or reorders patch lines. An LLM is given the compilation error, error types, and original diff, and is asked to return a strict subset together with an error-to-line mapping and removal rationale. The generated patch is first checked with \texttt{git apply --check}/\texttt{git apply} and then verified by running the build script in the failing build directory. If the refined patch fails to apply or compile, the pipeline retries once in a conservative mode; if it still fails, it falls back to the original developer patch. Only compile-verified outputs are marked as successful refinement and compilation. Recent-change clues are produced by extracting source files from the compiler error log, finding the previous commit for each file at the corresponding repository revision with \texttt{git log}, and retrieving the pre-failure diff with \texttt{git diff}; these clues are included in the sample's initial prompt text, so training and evaluation use the same injection path.

\subsection{Retention Rates of the Data-Construction Pipeline}
\label{app:data-pipeline-retention}

Table~\ref{tab:data-pipeline-retention} summarizes the per-stage retention from internal compilation-infrastructure logs to the final curated train/evaluation pool. Stage retention is computed relative to the previous stage, and cumulative retention is computed relative to the initial $12{,}618$ raw compilation-error directories. The historical-diff augmentation step adds context to each sample and does not further filter the sample pool.

\begin{table}[t]
    \centering
    \footnotesize
    \setlength{\tabcolsep}{4pt}
    \renewcommand{\arraystretch}{1.08}
    \caption{Per-stage sample retention in the data-construction pipeline.}
    \label{tab:data-pipeline-retention}
    \begin{tabular}{@{}p{0.23\linewidth} p{0.34\linewidth} p{0.11\linewidth} p{0.12\linewidth} p{0.12\linewidth}@{}}
        \toprule
        \textbf{Stage} & \textbf{Main processing} & \textbf{Retained samples} & \textbf{Stage retention} & \textbf{Cumulative retention} \\
        \midrule
        Raw compilation-error directories & collection from internal compilation infrastructure & $12{,}618$ & -- & $100.0\%$ \\
        \addlinespace[2pt]
        Candidate filtering & Multi-process remote patch extraction, 15+ error-type classification, deduplication, and token-length filtering & $6{,}190$ & $49.1\%$ & $49.1\%$ \\
        \addlinespace[2pt]
        Minimal-sufficient repair verification & LLM patch refinement and sandbox compilation-loop verification & $4{,}105$ & $66.3\%$ & $32.5\%$ \\
        \addlinespace[2pt]
        Context augmentation & Historical-diff context augmentation & $4{,}105$ & $100.0\%$ & $32.5\%$ \\
        \addlinespace[2pt]
        Final train/evaluation pool & Difficulty-aware filtering & $1{,}510$ & $36.8\%$ & $12.0\%$ \\
        \bottomrule
    \end{tabular}
\end{table}

\subsection{Semantic Judge Validation: 400-Sample Human Audit}
\label{app:judge-validation}

Both layered rewards and step-level process scores rely on the Kimi-K2.5 semantic judge. We treat it as a reward proxy and audit it offline; the audit does not change the training loop.

\paragraph{Protocol.}
The audit covers four checkpoints---GRPO-compile-only, GRPO-binary, GRPO-early (post-fork tail), and GRPO-full (final)---to characterize judge reliability \emph{across} methods and training stages, not only on the final main-method checkpoint. From each checkpoint's $400$-sample evaluation split, we stratify-sample $50$ judge-\texttt{consistent} and $50$ judge-\texttt{inconsistent} compile-success patches, for $100$ patches per checkpoint and $400$ patches in total. Compile-failure cases are excluded since the judge is not invoked. Two annotators familiar with the C/C++ build system independently label each patch as consistent or inconsistent with the GT minimal sufficient repair, blind to the judge's verdict and to the source checkpoint; a third annotator adjudicates disagreements.

\paragraph{Results.}
Table~\ref{tab:judge-validation} combines the pooled confusion matrix and aggregate reliability metrics. The two annotators agree on $372/400=0.930$ cases, and the judge agrees with the adjudicated human label on $360/400=0.900$ cases; precision/recall/$F_1$ on the \texttt{consistent} class are $0.880/0.917/0.898$. The slight precision--recall gap reflects a mild over-acceptance tendency typical of reference-conditioned LLM-as-judge on code. Per-checkpoint agreement is tightly clustered in $[89\%,91\%]$ (GRPO-compile-only: $89/100$, GRPO-binary: $90/100$, GRPO-early: $90/100$, GRPO-full: $91/100$), with no clear method-specific bias.

\begin{table}[t]
    \centering
    \footnotesize
    \setlength{\tabcolsep}{5pt}
    \renewcommand{\arraystretch}{1.05}
    \caption{Confusion matrix and reliability metrics for the $400$-sample human audit. Confidence intervals use Wilson for proportions and bootstrap ($10$k resamples) for $\kappa$.}
    \label{tab:judge-validation}
    \begin{tabular}{@{}l c c c@{}}
        \toprule
        \multicolumn{4}{@{}l}{\textbf{Confusion matrix pooled over four checkpoints}} \\
         & \textbf{Human: consistent} & \textbf{Human: inconsistent} & \textbf{Row total} \\
        \midrule
        \textbf{Judge: consistent}   & $176$ & $24$  & $200$ \\
        \textbf{Judge: inconsistent} & $16$  & $184$ & $200$ \\
        \textbf{Column total}        & $192$ & $208$ & $400$ \\
        \midrule
        \multicolumn{4}{@{}l}{\textbf{Aggregate metrics}} \\
        \textbf{Metric} & \textbf{Value} & \textbf{$95\%$ CI} & \textbf{Note} \\
        \midrule
        Inter-annotator agreement                         & $372/400=0.930$ & $[0.901, 0.951]$ & -- \\
        Inter-annotator $\kappa$                          & $0.86$          & $[0.81, 0.91]$   & -- \\
        Judge--human agreement                            & $360/400=0.900$ & $[0.867, 0.927]$ & -- \\
        Judge--human $\kappa$                             & $0.80$          & $[0.74, 0.86]$   & -- \\
        Precision / Recall / $F_1$ on \texttt{consistent} & $0.880/0.917/0.898$ & -- & -- \\
        \bottomrule
    \end{tabular}
\end{table}

\paragraph{Disagreement categories.}
Of the $40$ mismatches: \textbf{FP $=24$} (judge says consistent, human says inconsistent)---symptom-level fixes that suppress the diagnostic ($11$), unrelated behavior changes such as different fallback or extra logging ($7$), and coarser-granularity workarounds such as disabling a warning ($6$); \textbf{FN $=16$} (judge says inconsistent, human says consistent)---functionally equivalent repairs at different locations or surface forms ($10$), and alternate correct repair paths over-weighted as lexically divergent from GT ($6$). FP outnumbers FN by a moderate margin ($24$ vs.\ $16$), consistent with the well-documented over-acceptance bias of LLM-as-judge on code; the bias is distributed similarly across the four source checkpoints ($\chi^2$ test, $p=0.74$).

\paragraph{Supplementary note on the post-fork comparison.}
The judge is a bounded-noise reward proxy, not an oracle. The audit shows mild over-acceptance (FP $24$ vs.\ FN $16$), corresponding to an approximately $2.0$pp net offset in absolute success rate. Because GRPO-full and GRPO-early use the same judge, this offset mainly affects absolute levels; per-checkpoint judge--human agreement clusters in $[89\%,91\%]$, with no visible leniency toward GRPO-full over GRPO-early. Thus, the audit is only a supplementary check on the post-fork $5.0$pp increment, reducing the likelihood that it is entirely due to differential judge bias.

\paragraph{Paired significance and bias-corrected human-equivalent accuracy of the post-fork $+5.0$pp.}
Since GRPO-full and GRPO-early are evaluated on the same $400$-sample hint-free split, we run a paired McNemar test on \texttt{compile\_and\_semantic}: the contingency (both correct $176$, only-full $36$, only-early $16$, both wrong $172$) gives $\chi^2=(36-16)^2/(36+16)\approx7.69$ ($p\approx0.006$), with a Wilson $95\%$ CI of $[+0.018,+0.082]$ for the paired difference, excluding zero. Applying Rogan--Gladen judge-bias correction with the audit-derived sensitivity $0.917$ and specificity $0.885$ ($\hat p_{\text{true}}=(p_{\text{obs}}-(1-\text{spec}))/(\text{sens}+\text{spec}-1)$, denominator $0.802$), the human-equivalent rates are $0.455$ (GRPO-early) and $0.517$ (GRPO-full)---a corrected gap of $+6.2$pp, slightly larger than the raw $+5.0$pp; FP rates on \texttt{consistent}-judged patches for GRPO-full vs.\ GRPO-early ($5/50$ vs.\ $6/50$, $\chi^2=0.10$, $p=0.75$) further rule out method-specific judge leniency. The post-fork increment therefore survives both paired statistical inference and reasonable judge-bias correction.

\subsection{Dataset and Repository Scale}
\label{app:dataset-scale}

\begin{figure}[t]
    \centering
    \includegraphics[width=\linewidth]{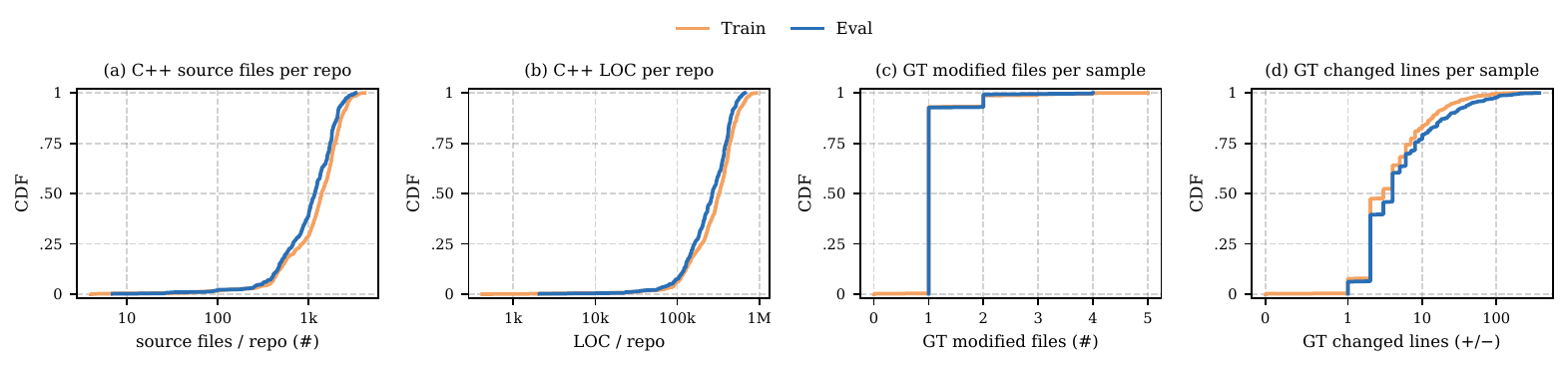}
    \caption{Distributions of repository scale and GT-modification scale across the $1{,}510$ samples (train $1{,}110$ + eval $400$). Each curve is the empirical CDF of one split; the $x$-axis is logarithmic in (a)(b)(d) and linear in (c). Source-file statistics include only C/C++ files and the companion \texttt{.proto} IDL, excluding build artifacts, logs, and configuration files.}
    \label{fig:dataset-scale}
\end{figure}

To characterize both the workload of the \emph{repair task} itself and the size of its \emph{repair context}, we measure four scale metrics per sample: (i) the repository-side \textbf{source file count}---the number of files in the repository subtree under C/C++ extensions (\texttt{.c/.cc/.cpp/.cxx}, the corresponding header families, and \texttt{.inc/.inl/.ipp}) plus the companion \texttt{.proto} IDL files; (ii) the repository-side \textbf{source LOC}---the total line count of those files; (iii) the \textbf{number of GT-modified files}; and (iv) the \textbf{number of GT-changed lines}---i.e., the modified-file count and the combined added/deleted line count of the minimal sufficient repair patch. Figure~\ref{fig:dataset-scale} reports the empirical CDFs of these four metrics on the train and eval splits.

\paragraph{Strong asymmetry between read and write scale.}
Across the $1{,}510$ samples, the per-repository C++ source-file count has median $1{,}339$ and maximum $4{,}135$, while source LOC has median ${\sim}3.08\!\times\!10^5$ and maximum ${\sim}9.18\!\times\!10^5$ (Figure~\ref{fig:dataset-scale}(a)(b)). In contrast, the number of GT-modified files has median $1$, $p_{75}\!=\!1$, and maximum $5$; the total GT-changed line count has median only $3$, $p_{75}\!=\!7$, and maximum $378$ (Figure~\ref{fig:dataset-scale}(c)(d)). The policy must therefore localize $O(1)$ files and $O(10)$ lines of precise edits within a context of $10^3$-scale files and $10^5$-scale LOC---the \emph{read scale exceeds the write scale by roughly $10^4$ to $10^5$ times}.

\paragraph{Train/eval alignment.}
The train and eval curves in Figure~\ref{fig:dataset-scale} overlap on all four metrics: source-file count and LOC have similar medians (train: $1{,}393$ files / $3.26\!\times\!10^5$ LOC; eval: $1{,}172$ files / $2.68\!\times\!10^5$ LOC), and only the right tail of the train split extends slightly further toward larger repositories (max $4{,}135$ files / $9.18\!\times\!10^5$ LOC vs. eval's $3{,}300$ files / $6.68\!\times\!10^5$ LOC). The GT-modification scale is essentially identical between splits, supporting distributional comparability of fixed-split evaluation metrics.

\subsection{GRPO Training Hyperparameters and Sampling Budget}
\label{app:grpo-hparams}

Table~\ref{tab:grpo-hparams} summarizes the training, rollout, and optimizer configuration shared by the main experiments and ablations. The optimizer follows AdamW~\citep{loshchilov2019adamw} with decoupled weight decay.

\begin{table}[t]
    \centering
    \footnotesize
    \setlength{\tabcolsep}{4pt}
    \renewcommand{\arraystretch}{0.96}
    \caption{GRPO training and sampling hyperparameters used across experiments.}
    \label{tab:grpo-hparams}
    \begin{tabular}{@{}p{0.38\linewidth} p{0.56\linewidth}@{}}
        \toprule
        \textbf{Hyperparameter} & \textbf{Value} \\
        \midrule
        \multicolumn{2}{@{}l}{\textbf{Base model and training system}} \\
        Base model              & Qwen3-Coder-30B-A3B-Instruct \\
        Training backend         & Fully Sharded Data Parallel (FSDP) \\
        Attention                & \texttt{flash\_attention\_3} \\
        \texttt{rotary-base}     & $10^{7}$ \\
        \midrule
        \multicolumn{2}{@{}l}{\textbf{Sampling}} \\
        \texttt{rollout-temperature}       & $1$ \\
        \texttt{top-p}                     & $1$ \\
        \texttt{rollout-max-context-len}   & $40$k \\
        \texttt{rollout-max-response-len}  & $8$k \\
        \texttt{n-samples-per-prompt}      & $8$ \\
        \texttt{rollout-batch-size}        & $8$ prompts \\
        \texttt{global-batch-size}         & $64$ \\
        \midrule
        \multicolumn{2}{@{}l}{\textbf{GRPO loss}} \\
        \texttt{advantage-estimator} & \texttt{grpo} \\
        \texttt{eps-clip} (low)      & $0.2$ \\
        \texttt{eps-clip-high}       & $0.28$ \\
        \texttt{use-kl-loss}         & \texttt{low\_var\_kl} \\
        \texttt{kl-loss-coef}        & $0.01$ \\
        \texttt{entropy-coef}        & $0$ \\
        \midrule
        \multicolumn{2}{@{}l}{\textbf{Optimizer}} \\
        Optimizer              & AdamW \\
        \texttt{lr}           & $3\times10^{-6}$ (constant) \\
        $\beta_1$, $\beta_2$  & $0.9$, $0.98$ \\
        \texttt{weight-decay} & $0.1$ \\
        \midrule
        \multicolumn{2}{@{}l}{\textbf{Evaluation}} \\
        \texttt{eval-interval}         & $20$ \\
        \texttt{eval-max-response-len} & $40$k \\
        \texttt{eval-top-p}           & $1$ \\
        Evaluation set                 & fixed split of $400$ samples \\
        \midrule
        \multicolumn{2}{@{}l}{\textbf{Trajectory control}} \\
        \texttt{max\_steps}            & $50$ \\
        Compile-timeout abort threshold & $2$ \\
        \bottomrule
    \end{tabular}
\end{table}

\subsection{Rollout Governance Implementation Details}
\label{app:rollout-impl}

\paragraph{Three-way concurrency control used in training.}
The rollout runtime separately limits three resource classes: LLM inference, ordinary sandbox operations, and compilation. Sandbox-side I/O and reward computation are scheduled asynchronously, preventing long tool calls from blocking the rollout controller.
\begin{itemize}
    \item \emph{Resource mapping}: \texttt{compile\_code} calls and the final compilation check at trajectory termination use the compilation resource pool; file editing, command execution, and environment reset use the sandbox resource pool.
    \item \emph{Inference-slot release}: inference resources cover only a single model-generation request and do not cover subsequent tool execution, so tool I/O can overlap with other rollout generations.
    \item \emph{Compilation concurrency constraint}: compilation tasks are queued under a conservative concurrency limit to prevent CPU, memory, and I/O contention from contaminating the execution opportunity of rollouts in the same group.
    \item \emph{Cascading protection for compilation timeouts}: if a single trajectory experiences at least the compile-timeout abort threshold (default $2$) consecutive compilation timeouts, it is proactively terminated at the next step and marked as \texttt{consecutive\_}\allowbreak\texttt{compile\_}\allowbreak\texttt{timeouts}. This prevents a small number of toxic samples from continuously occupying compilation channels and slowing down the entire rollout batch.
\end{itemize}

\paragraph{Exception classification and handling.}
\texttt{abnormal\_exit\_type} is drawn from \{\texttt{context\_limit}, \texttt{abort}, \texttt{max\_tokens}, \texttt{catastrophic\_repetition}, \texttt{excessive\_tool\_calls}, \texttt{consecutive\_}\allowbreak\texttt{compile\_}\allowbreak\texttt{timeouts}\}. The handling rules are:
\begin{itemize}
    \item \emph{Full-trajectory masking}: \texttt{context\_limit}, \texttt{abort}, \texttt{max\_tokens}, \texttt{consecutive\_}\allowbreak\texttt{compile\_}\allowbreak\texttt{timeouts}, and \texttt{environment\_}\allowbreak\texttt{setup\_failed}. Final compilation and semantic judgment are skipped to save evaluation budget.
    \item \emph{Last-step retention}: \texttt{catastrophic\_repetition} and \texttt{excessive\_tool\_calls}, implemented as \texttt{mask\_except\_last\_step}.
    \item \emph{Normal settlement}: \texttt{finish\_called}, \texttt{max\_steps}, and \texttt{no\_tool\_call\_stop}.
\end{itemize}
The experiments use the following trigger rules: \texttt{context\_limit} fires when prompt plus response tokens reach \texttt{rollout-max-context-len}; \texttt{max\_tokens} fires when a single generation step hits \texttt{rollout-max-response-len}; \texttt{excessive\_tool\_calls} fires when one assistant turn contains more than $5$ parsed tool calls, and only that turn's model tokens are retained as the penalty sample; \texttt{consecutive\_}\allowbreak\texttt{compile\_}\allowbreak\texttt{timeouts} fires after $2$ consecutive \texttt{compile\_code} timeouts in the same trajectory; and \texttt{max\_steps} is treated as normal budget exhaustion with the default budget of $50$ steps. \texttt{catastrophic\_repetition} fires when a $15$--$50$-token n-gram repeats more than $30$ times; once it fires, the trajectory is terminated early and only the final assistant turn's tokens are retained as a penalty sample.

\paragraph{Reward and process-score implementation.}
The main training reward emits layered outputs $\{0, 0.5, 1\}$, where $0$ means no successful compilation, $0.5$ means successful compilation with semantic inconsistency, skipped semantic checking, or semantic-judge failure, and $1$ means successful compilation with semantic consistency. In addition, Kimi-K2.5 assigns a step-level process score $s_i\in[0,1]$ to each tool-call step. This score is broadcast to all assistant tokens in the corresponding step and multiplied by GRPO's own loss mask to scale the update magnitude of model-generated tokens by step. Process scores are applied only when compilation succeeds and process scoring returns valid scores; ordinary non-compiling failures receive no step-level reweighting and therefore keep the original assistant-only mask (equivalent to $s_i\equiv1$). Abnormal samples are still retained in the same-group reward normalization to keep the rollout group size $K$ fixed: full-trajectory-masked samples receive $R=0$ and a zero loss mask, so they affect only the group mean/variance and produce no gradient; last-step-retention samples also receive $R=0$, but retain only the final assistant turn as the penalty sample. When process scores are disabled, $s_i\equiv1$, yielding the layered baseline used to isolate the effect of process-level credit assignment. The binary reward ablation separately collapses the layered reward output into $\{0,1\}$ to test outcome-level reward semantics loss.

\section{Failure Analysis of Token-Level Policy Distillation}
\label{app:distill}

\subsection{Research Position on Distilling Privileged Information}
\label{app:distill_positioning}

After obtaining a stably trainable GRPO baseline, we ask whether privileged hints $I$, visible during training but strictly unavailable during evaluation, can provide a token-level dense signal that replaces the process-credit signal from step-level process scores, without changing the test-time interface. We instantiate this path with token-level OPD as defined in Eq.~\eqref{eq:pi-distill}--\eqref{eq:opsd} of Appendix~\ref{app:distill-impl}. Our answer is \textbf{no}. At the objective level, the token-level KL in Eq.~\eqref{eq:pi-distill}--\eqref{eq:opsd} optimizes local distributional alignment with the teacher, whereas GRPO in Eq.~\eqref{eq:grpo-adv}--\eqref{eq:grpo-loss} optimizes outcome ranking among trajectories from the same prompt. In weak-feedback, long-trajectory code-agent tasks, success or failure is determined by a small number of tool calls, edit submissions, and termination decisions. Token-level KL therefore dilutes supervision over many non-critical tokens, violating one of the implicit assumptions discussed in Section~\ref{sec:prelim_opd}; after KL is injected into $R'_{\text{env}}$, it also changes the within-group return scale used by GRPO. Consequently, token-level distillation cannot efficiently replace the process-credit signal provided by step-level process scores.

Based on this view, we evaluate OPSD and $\pi$-Distill under the same rollout pipeline, layered reward, and strictly hint-free evaluation protocol, using them as comparison methods for the token-level supervision path. This subsection states the research position and conclusion; the following sections analyze failure mechanisms and training diagnostics.

\subsection{Negative Evidence: Token-Level Policy Distillation Does Not Replace Process Credit}
\label{app:distill_failure}

\paragraph{Experimental setup.}
The loss forms and implementation details of the two methods are given in Eq.~\eqref{eq:pi-distill}--\eqref{eq:opsd} of Appendix~\ref{app:distill-impl}. Both OPSD and $\pi$-Distill are \emph{stacked on top of GRPO-early}: they are initialized from the GRPO-early fork checkpoint and compared with the GRPO-early post-fork baseline over the shared \texttt{rollout/step}\,$\in (239, 399]$ window; apart from the distillation-specific token-level KL term (written into $\widetilde{R}$ as in Eq.~\eqref{eq:pi-distill-reward}/\eqref{eq:opsd-reward}, with the teacher branch reading prompts augmented by the \texttt{<Privileged hint>} block), every other setting is identical to GRPO-early.

\begin{figure}[t]
    \centering
    \includegraphics[width=\linewidth]{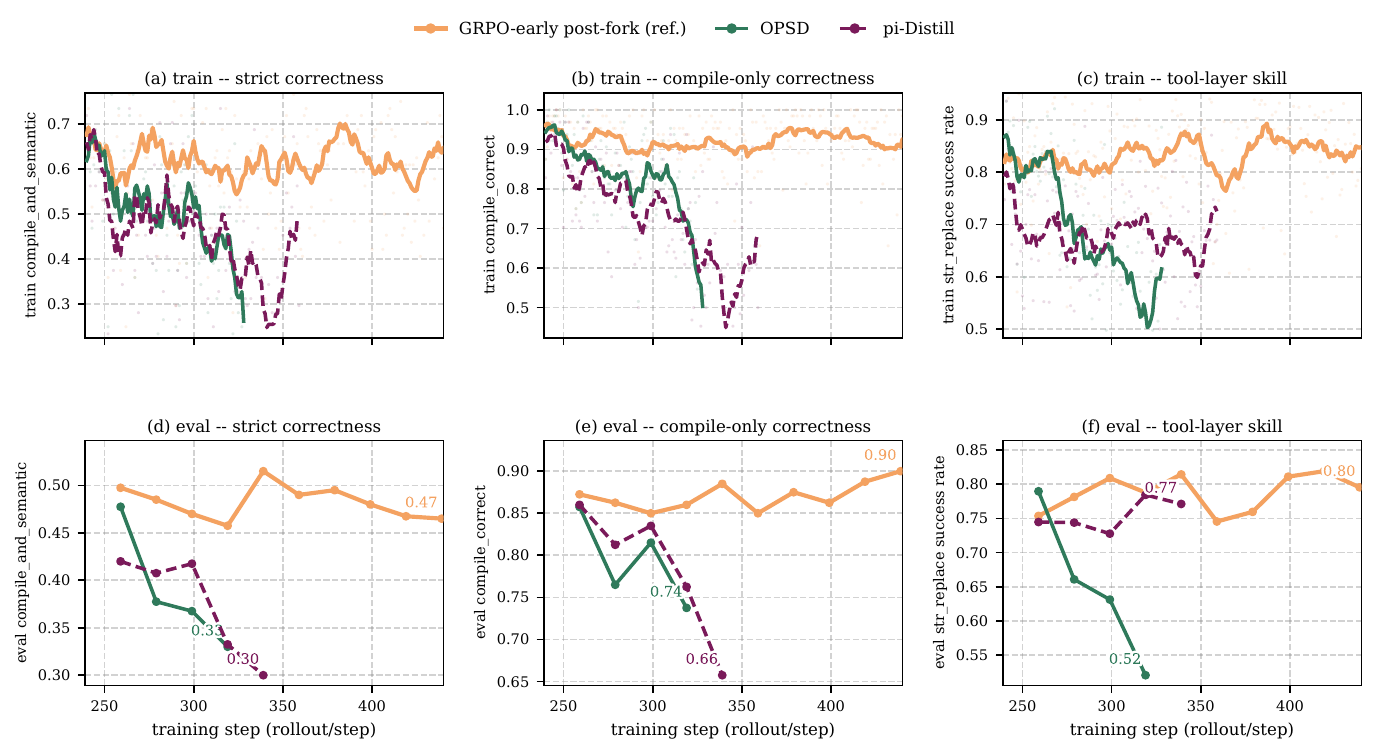}
    \caption{Outcome-level comparison of distillation over the shared \texttt{rollout/step}\,$\in (239, 399]$ window. The top and bottom rows show correctness, tool-level success, and exit-behavior metrics on training rollouts (on-policy) and strictly hint-free evaluation, respectively.}
    \label{fig:distill_outcomes}
\end{figure}

\paragraph{Outcome level: correctness lags across the board.}
The top row (train) and bottom row (eval) of Figure~\ref{fig:distill_outcomes} show a central fact: regardless of KL direction, token-level distillation yields substantially lower strict outcome correctness than the GRPO-early post-fork baseline, and the gap shows no sign of closing throughout the shared window.

Specifically, the strict correctness rate on evaluation, \texttt{compile\_and\_semantic}, is only about $0.36$ for OPSD and $0.35$ for $\pi$-Distill, whereas GRPO-early post-fork stays around $0.48$ in the same window. Compilation correctness, \texttt{compile\_correct}, also lags behind, with OPSD at approximately $0.77$, $\pi$-Distill at approximately $0.75$, and the baseline at approximately $0.88$. This indicates that the failure is not evaluation noise, but a real gap in policy quality.

The more informative divergence appears at the tool and behavior levels. The evaluation \texttt{str\_replace} success rate of $\pi$-Distill, approximately $0.76$, is close to GRPO-early, approximately $0.81$, suggesting that its tool-use skill is largely preserved. However, its evaluation average steps are only about $12$, much lower than the baseline, approximately $21$, and OPSD, approximately $23$. This means that the $\pi$-Distill agent learns to take actions and even to make plausible edits, but does not learn to terminate at the right time: it calls \texttt{finish} prematurely and submits incomplete repairs, causing both strict correctness and compilation correctness to drop. In contrast, OPSD also shows a clear degradation in \texttt{str\_replace} success rate, approximately $0.60$, indicating that reverse-KL-induced low entropy not only harms termination decisions but also erodes tool-level skill.

Overall, the failure of token-level distillation is not a single-dimensional inability to learn. $\pi$-Distill preserves local skill while losing global pacing, manifested as premature termination; OPSD loses both tool precision and termination timing while compressing distributional entropy. Neither path yields a substantive improvement in strict outcome correctness.

\begin{figure}[t]
    \centering
    \includegraphics[width=\linewidth]{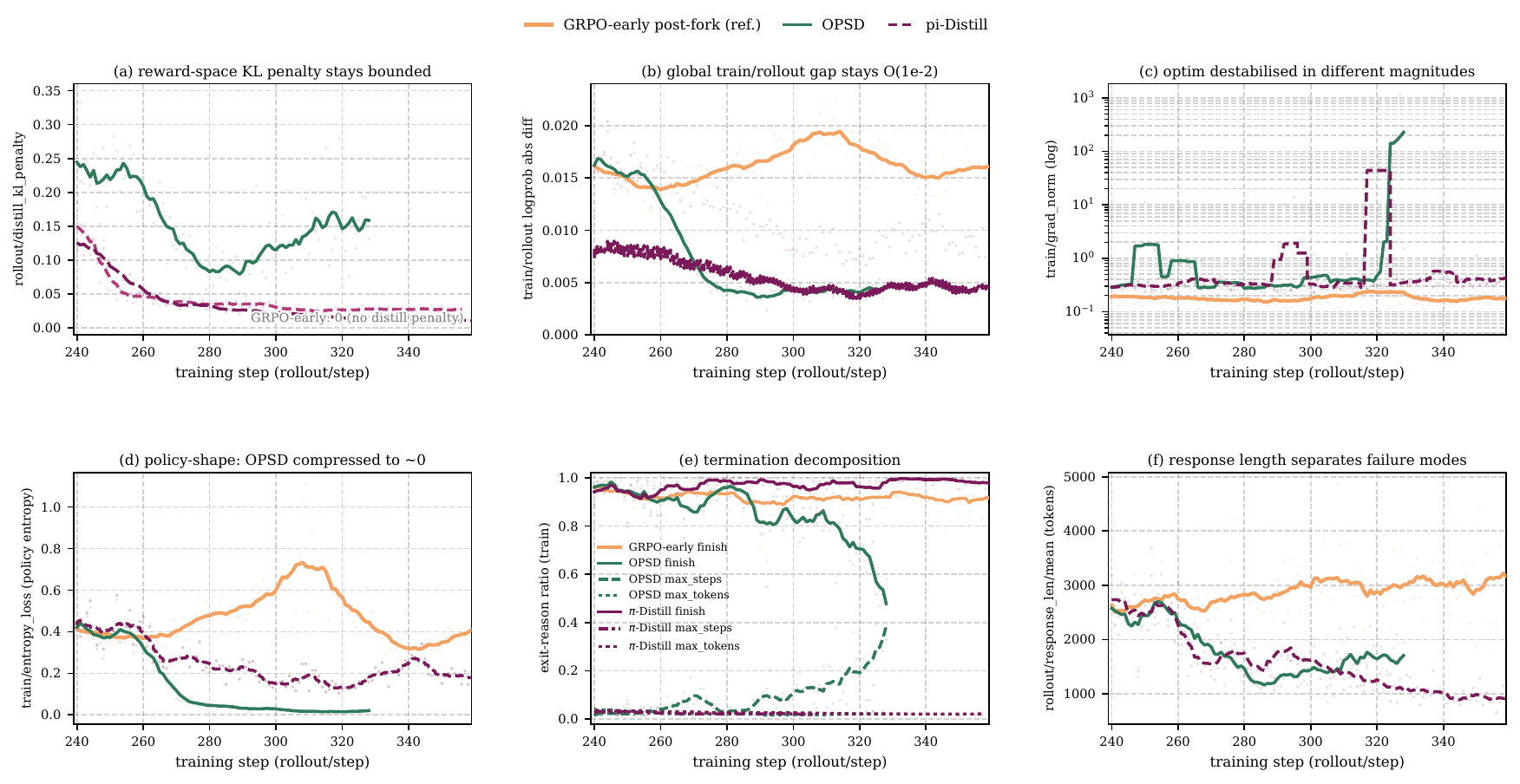}
    \caption{Mechanism-level diagnostics for distillation over the same shared window. The figure shows optimization and behavioral dynamics, including reward-space KL penalty, log-probability gap, gradient norm, policy entropy, and exit reasons.}
    \label{fig:distill_mechanism}
\end{figure}

\paragraph{Mechanism level: progressive evidence from six panels.}
Figure~\ref{fig:distill_mechanism} provides mechanistic evidence along six dimensions: KL, global mismatch, optimizer behavior, policy shape, exit structure, and length drift. The evidence can be summarized in three layers.

The first layer rules out simple explanations. In panel (a), the reward-space distillation penalty is bounded for both methods and generally decreases, so the failure cannot be attributed to an exploding KL penalty. In panel (b), the estimator-agnostic train-rollout log-probability gap remains at the $O(10^{-2})$ scale for all three runs. $\pi$-Distill does not clearly separate from the baseline on this global mismatch scalar, so a single global policy drift cannot explain the failure either.

The second layer shows divergence in optimization and policy shape. In panel (c), \texttt{grad\_norm} on a log scale reaches approximately $10^2$ for OPSD, $10^0$ for $\pi$-Distill, and $10^{-1}$ for the baseline. Both distillation methods perturb optimization stability, but through different entry points and magnitudes. Panel (d) shows that OPSD compresses policy entropy nearly to zero, whereas $\pi$-Distill maintains substantially higher entropy. This indicates that OPSD anchors the policy to a narrow teacher-induced distribution, while $\pi$-Distill does not undergo comparable compression.

The third layer identifies behavioral divergence. Panel (e) shows that OPSD has decreasing \texttt{finish\_called}, increasing \texttt{max\_steps}, and nearly unchanged \texttt{max\_tokens}; $\pi$-Distill exhibits a different pattern, with \texttt{finish\_called} as high as approximately $0.98$ and both \texttt{max\_steps} and \texttt{max\_tokens} compressed close to zero. Panel (f) shows that response length remains relatively short for OPSD, while $\pi$-Distill contracts sharply to approximately $952$ tokens, even shorter than the GRPO-early baseline. Combining panels (d), (e), and (f), OPSD corresponds to a low-entropy form of failure in which the policy becomes reluctant or unable to terminate appropriately, with entropy near zero, unchanged \texttt{max\_tokens}, and no length expansion. $\pi$-Distill instead corresponds to premature termination: length is controlled, entropy remains open, and gradient norm is moderate, but evaluation strict correctness still does not improve.

These three layers show that both methods inject token-level supervision along long trajectories. OPSD drives the policy toward low entropy and degraded termination behavior, whereas $\pi$-Distill removes length expansion and gradient instability at the mechanism level but moves toward a subtler local optimum: premature termination.

\paragraph{Root-cause explanation.}
Combining outcome-level and mechanism-level diagnostics, the failure can be attributed to one common root cause and two method-specific paths.

\textbf{(A) Common root cause: mismatch between supervision granularity and decision granularity.} GRPO optimizes the relative outcome ranking of complete trajectories under the same prompt, while token-level distillation constrains local distributional alignment at each token. For long-trajectory, weak-feedback tasks such as compile-fix, success is often determined by a small number of action-level branching points, especially whether to continue using tools or call \texttt{finish}. In contrast, many schema, path, log, and NL tokens are nearly neutral with respect to the final reward. Token-level supervision is therefore easily diluted by many low-information positions. Moreover, the implementation uses \texttt{masked sum} to aggregate KL, so distillation pressure accumulates with sequence length and couples with statistics of non-critical tokens. The teacher signal is not concentrated on key decisions; instead, it first reshapes local distributions and behavioral rhythm over long trajectories.

The direct consequence of this mismatch is degraded policy quality: both distillation methods maintain lower \texttt{str\_replace} success and \texttt{compile\_and\_semantic} rates than the GRPO baseline. On top of this shared failure, the different KL directions further separate the degradation paths. OPSD's reverse-KL penalty favors narrow-band alignment; after masked-sum aggregation, it changes trajectory-level rewards and amplifies the corresponding policy-gradient updates, manifested as low entropy and degraded termination behavior. $\pi$-Distill's forward-KL penalty is more tolerant of distributional support; panel (a) shows that its KL penalty is bounded and even lower than OPSD's. However, token-level supervision is diluted by many neutral tokens, and the policy drifts in an open-entropy space toward a surface shortcut: very high \texttt{finish\_called} with low outcome correctness. We detail the two paths below.

\textbf{(B) Method-specific path of $\pi$-Distill: mode covering under forward KL and supervision dilution lead to premature termination.} $\pi$-Distill uses $\mathrm{KL}(\pi_T\|\pi_S)$, whose mode-covering property requires the student to cover the full probability support of the teacher and therefore does not compress the distribution into a single narrow band. Panel (a) shows that its distillation penalty is bounded and even below OPSD's; entropy remains open, gradient norm is moderate, and training remains stable. However, the granularity mismatch of token-level supervision remains. As a reward penalty, the forward-KL form encourages coverage of teacher support rather than concentrating trajectory-level advantages on the few high-quality repair decisions. The policy therefore drifts in an open-entropy space toward a surface shortcut. The teacher hint contains many language patterns of summarization and submission; by aligning to these patterns, the policy learns the rhythm of finishing early, but not the causal logic of submitting only after the repair is complete. As a result, \texttt{finish\_called} reaches approximately $0.98$, while evaluation strict correctness remains only about $0.35$, substantially below GRPO-early's approximately $0.48$.

\textbf{(C) Method-specific path of OPSD: mode seeking under reverse KL and long-trajectory gradient accumulation cause low entropy and gradient amplification.} OPSD uses $\mathrm{KL}(\pi_S\|\pi_T)$. Its mode-seeking property directly pushes the policy toward the narrow band of the teacher distribution, thereby suppressing exploration. In weak-feedback, long-trajectory tasks, this rigid alignment does not necessarily prioritize the few key edits needed for final correctness; instead, it first compresses overall distributional entropy and termination choices. The evidence in Figure~\ref{fig:distill_mechanism} shows three aspects of OPSD's failure: panel (d) shows that policy entropy is compressed nearly to zero, meaning that the distribution is anchored to a narrow teacher-induced band; panel (c) shows that \texttt{grad\_norm} grows to approximately $10^2$, indicating that the reverse-KL penalty, after being written into trajectory rewards through masked-sum aggregation, changes within-group advantages and substantially amplifies policy-gradient updates over long trajectories. The mechanism here is not direct backpropagation through the KL term itself; it is the effect of KL-shaped rewards on GRPO's relative reward scale. Compared with $\pi$-Distill, the narrow-band pressure of reverse KL more readily directs these advantages toward entropy collapse. Panels (e) and (f) show that \texttt{max\_tokens} barely changes and response length does not expand, indicating that pressure is not expressed as length growth but as low entropy and termination degradation. Behaviorally, this appears as degraded \texttt{finish} decisions, making trajectories harder to end at the appropriate time.

\paragraph{Conclusion.}
In this setting, token-level distillation does not replace the process-credit signal provided by step-level process scores: it does not translate into better outcome correctness in evaluation, nor does it solve the attribution of credit to key steps. It also does not itself repair outcome-level reward semantics or within-group comparability, and introduces additional optimization issues. $\pi$-Distill further shows that even when KL strength is comparable to OPSD and mechanism-level indicators appear normal, with no length expansion or gradient failure, evaluation strict correctness still does not match the GRPO baseline. The limitation of token-level alignment is therefore not determined by an overly strong KL alone, but by the mismatch between supervision granularity and decision granularity. This result is consistent with the main conclusions from the previous sections: layered rewards and step-level process-score weighting are more directly matched to the task than token-level alignment. This conclusion is also consistent with recent reports on the failure of token-level OPD under distribution shift~\citep{fu2026revisiting}, and suggests that future work should explore tool-call-level supervision, trajectory-level counterfactual correction~\citep{lyu2025score}, or teacher-consistency gating~\citep{stein2026gates}.

\subsection{On-Policy Distillation Implementation Details}
\label{app:distill-impl}

\paragraph{$\pi$-Distill.}
Let $\pi^T_\theta(\cdot|s,I)$ denote the teacher with privileged information, and let $\pi^S_\theta(\cdot|s)$ denote the student without privileged information; the two share parameters. The standard $\pi$-Distill objective contains both teacher and student terms:
\begin{align}
J_{\text{Teacher}}(\theta)
&= \mathbb{E}_{o\sim\pi^T_\theta(\cdot|s,I),\,s\sim P}\!\bigl[R(o,s)\bigr]
-\beta D_{\mathrm{KL}}\!\left(\pi^T_\theta(o|s,I)\,\|\,\mathrm{sg}\bigl(\pi^S_\theta(o|s)\bigr)\right), \\
J_{\text{Student}}(\theta)
&= \mathbb{E}_{o\sim\pi^T_\theta(\cdot|s,I),\,s\sim P}\!\left[
\frac{\pi^S_\theta(o|s)}{\mathrm{sg}\bigl(\pi^T_\theta(o|s,I)\bigr)}R(o,s)
\right]
-\beta D_{\mathrm{KL}}\!\left(\mathrm{sg}\bigl(\pi^T_\theta(o|s,I)\bigr)\,\|\,\pi^S_\theta(o|s)\right), \\
J_{\pi\text{-Distill}}(\theta)
&= \alpha J_{\text{Teacher}}(\theta)+(1-\alpha)J_{\text{Student}}(\theta).
\label{eq:pi-distill}
\end{align}
In the compile-fix implementation, the KL regularizers in the joint objective above are merged into a single term and applied as a reward-shaping penalty entering the GRPO group advantage. The teacher samples trajectories on the hint-augmented prompt, and the reference policy is taken to be the stop-gradient distribution of the current hint-free student branch:
\begin{equation}
\widetilde{R}_{\pi\text{-Distill}}(o_k,s)
= R_{\text{env}}(o_k,s)
- \beta\,D^{\text{masked-sum}}_{\mathrm{KL}}\!\left(
\pi^T_\theta(\cdot|s,I)\,\|\,\mathrm{sg}\bigl(\pi^S_\theta(\cdot|s)\bigr)
\right).
\label{eq:pi-distill-reward}
\end{equation}
Within-group advantages are then computed from $\widetilde{R}_{\pi\text{-Distill}}$. The implementation uses \emph{batch expansion} to retain both a student-copy and a teacher-copy, which share the same $\widetilde{R}_{\pi\text{-Distill}}$ and the same group advantage and differ only in the importance ratio: the student-copy uses hint-free tokens with $\rho^S_k=\pi^S_\theta(o_k|s)/\pi^T_{\mathrm{old}}(o_k|s,I)$ for off-policy GRPO, while the teacher-copy uses hint-augmented tokens with $\rho^T_k=\pi^T_\theta(o_k|s,I)/\pi^T_{\mathrm{old}}(o_k|s,I)$; the two copies have loss weights $2(1-\alpha)$ and $2\alpha$ so that the average over the doubled batch reproduces $\alpha J_{\text{Teacher}}+(1-\alpha)J_{\text{Student}}$, with default $\alpha=0.5$. The KL is estimated with \texttt{masked-sum} / RB top-$k$ over assistant-generation tokens only; defaults are $\beta=0.01$, RB top-$k=32$, and C3 hints.

\paragraph{OPSD.}
OPSD uses the student policy as the sampling policy: the hint-free student performs on-policy rollout, while the teacher is the same shared-parameter model conditioned on a hint-augmented prompt and scores the same trajectory. Its objective is
\begin{equation}
J_{\mathrm{OPSD}}(\theta)
= \mathbb{E}_{o\sim\pi^S_\theta(\cdot|s),\,s\sim P}\!\bigl[R(o,s)\bigr]
-\beta D_{\mathrm{KL}}\!\left(
\pi^S_\theta(o|s)\,\|\,\mathrm{sg}\bigl(\pi^T_\theta(o|s,I)\bigr)
\right),
\label{eq:opsd-objective}
\end{equation}
where the reverse KL can be viewed as a dense token-level reward measuring how closely the hint-free student matches the hint-conditioned teacher. In the implementation, this KL term is first written into the raw reward:
\begin{equation}
\widetilde{R}_{\mathrm{OPSD}}(o_{i,k},s_i)
= R_{\text{env}}(o_{i,k},s_i)
- \beta\,D^{\text{masked-sum}}_{\mathrm{KL}}\!\left(
\pi^S_\theta(\cdot|s_i)\,\|\,\mathrm{sg}\bigl(\pi^T_\theta(\cdot|s_i,I_i)\bigr)
\right),
\label{eq:opsd-reward}
\end{equation}
and then used to compute group-normalized advantages:
\begin{equation}
\hat A_{i,k}=
\frac{
\widetilde{R}_{\mathrm{OPSD}}(o_{i,k},s_i)
-\frac{1}{K}\sum_{j=1}^{K}\widetilde{R}_{\mathrm{OPSD}}(o_{i,j},s_i)}
{\mathrm{std}(\{\widetilde{R}_{\mathrm{OPSD}}(o_{i,j},s_i)\}_{j=1}^{K})},
\qquad
\rho_{i,k}=\frac{\pi^S_\theta(o_{i,k}|s_i)}{\pi^S_{\mathrm{old}}(o_{i,k}|s_i)}.
\label{eq:opsd}
\end{equation}
In practice, this group-normalized advantage is implemented by slime's GRPO reward normalization. Teacher log-probabilities are precomputed during rollout and used with stop-gradient; KL is estimated with \texttt{masked-sum} / RB top-$k$ over assistant-generation tokens only. The defaults are $\beta=0.02$, RB top-$k=32$, and C3 hints.

\paragraph{Implementation details.}
Both $\pi$-Distill and OPSD subtract a KL penalty from the raw reward before computing GRPO within-group advantages, with KL directions $D_{\mathrm{KL}}(\pi^T\|\pi^S)$ and $D_{\mathrm{KL}}(\pi^S\|\pi^T)$, respectively. To adapt to the long trajectories and tool interactions in compile-fix, we estimate sequence-level KL only on assistant-generation tokens covered by the loss mask and use a \texttt{masked-sum} / RB top-$k=32$ approximation; $\beta$ is scaled to $\{0.01,0.02\}$; privileged information is fixed to C3 hints abstracted from GT diffs.

\subsection{Full OPSD Sweep and Additional Diagnostics}
\label{app:opsd-full}

Following the presentation in Appendix~\ref{app:distill_failure}, the main paper keeps only one representative OPSD run that enters long-horizon comparison, while early-stopped experiments and other sweep configurations are reported uniformly in the appendix. This section summarizes the OPSD training configurations. The goal is not to compare every setting in the main text, but to show common patterns under different KL injection forms and strengths. The diagnostics are as follows:
\begin{itemize}
    \item per-token KL in advantage ($\beta\!=\!1$): the gradient norm increases rapidly at the beginning of training, and the loss soon diverges.
    \item masked mean KL in reward ($\beta\!=\!0.25$): the KL term remains very small for a long period and has limited effect on group-normalized relative returns; the training trajectory nearly overlaps with the pure GRPO baseline, with no stable gain observed.
    \item masked sum KL in reward ($\beta\!=\!0.25$): the KL contribution fluctuates substantially with trajectory length, so the same $\beta$ corresponds to different penalty strengths for short and long samples; training becomes clearly unstable and terminates early.
    \item masked sum KL in reward ($\beta\!=\!0.05$): this setting also exhibits length-coupled optimization instability. Although it is slightly milder than the larger coefficient, it still does not enter a comparable long-horizon interval.
    \item masked sum KL in reward ($\beta\!=\!0.02$): this is the only masked-sum configuration that enters long-horizon comparison. Its explicit KL can be continuously reduced, but correctness evaluation does not outperform GRPO-early post-fork, and the run gradually shows termination-behavior degradation and low entropy. It therefore serves as a runnable but non-beneficial counterexample, rather than positive evidence that OPSD can replace GRPO's signal conditions.
    \item masked mean KL in reward ($\beta\!=\!50$): training can formally continue, but learning is clearly suppressed, eventually producing long-term reward degradation and interruption in the later stage.
\end{itemize}
These additional diagnostics are consistent with the failure-mechanism analysis in the main text. On the one hand, token-level KL over long trajectories easily couples with length and non-critical token statistics. On the other hand, even when the explicit KL is reduced to a low range, we do not identify an OPSD operating regime that simultaneously maintains training stability, normal behavior, and improved outcome correctness. In other words, larger KL coefficients more directly induce instability, while smaller coefficients may allow training to continue for longer but still mainly produce termination-behavior degradation or suppressed learning rather than correctness improvement.

\section{Core Prompt Design}
\label{app:prompts}

This section lists the main prompt templates used in the experiments. All prompts are iteratively tuned to ensure stable output format and consistent judgment criteria. The templates below preserve placeholders, control tags, and JavaScript Object Notation (JSON) fields.

\subsection{System Prompt: Role Definition and Tool Instructions}
\label{app:prompt-system}

The system prompt is injected through the tokenizer chat-template API as the top-level role constraint for each interaction round of the agent. The template is shown below.

\begin{verbatim}
You are an experienced C++ engineer specializing in fixing compilation errors.

## Task Description
Locate and fix code issues based on compiler error messages.

## Repair Requirements
1. Fix only the code that causes compilation ERRORs; do not fix WARNINGs.
2. Ignore all compilation warnings and focus only on compilation errors.
3. Do not break the original logic or functionality of the code; remain consistent
   with the original intent.
4. Make only necessary changes and do not introduce new issues.
5. Ensure that the repaired code compiles successfully. Warnings are allowed, but
   errors are not.

## Workflow
1. **Analyze errors**: Carefully read the compiler error messages and understand
   the specific cause of each error.
2. **Make a plan**: Use the plan command of task_tracker to create a task list.
3. **Locate the issue**: Use the view command of str_replace_editor to inspect
   the reported file.
4. **Understand context**: When necessary, use bash to search for relevant
   definitions, declarations, or usages.
5. **Make a precise fix**: Use str_replace in str_replace_editor to make a
   minimal modification.
6. **Update progress**: After completing each repair, update the task status with
   task_tracker.
7. **Verify the result**: Call compile_code to recompile.
8. **Iterate**: If errors remain, repeat steps 3--7.
9. **Submit**: After confirming successful compilation, call finish to submit.

## Available Tools
- task_tracker: task management (plan / view)
- str_replace_editor: file operations (view / str_replace / create / insert)
- bash: execute shell commands
- compile_code: compile and verify
- finish: submit after the repair is complete

## Notes
- Make precise changes and avoid introducing new issues.
- Call compile_code and confirm success before calling finish.
- When calling finish, briefly explain the cause of the issue and the solution.
\end{verbatim}

\subsection{User Task Prompt: Initial Observation}
\label{app:prompt-user}

During data preprocessing, the raw compilation error log is wrapped as the following user message, which serves as the initial observation of the trajectory. If a teacher hint is enabled, a \texttt{<Privileged hint>} block in the format of Section~\ref{app:prompt-hint} is inserted at the beginning of the user message.

\begin{verbatim}
# Compilation Error Repair Task

## Problem Overview
The code fails to compile. Analyze the error log and fix the issue.

## Compilation Error Log
```
{error_log}
```

## Task Requirements
1. Analyze the compilation errors above and identify the root cause.
2. Propose a minimal code modification.
3. Ensure that the repaired code compiles successfully.

Please start analyzing and fixing the compilation error.
\end{verbatim}

\subsection{Semantic-Consistency Judgment Prompt: Terminal Reward \texorpdfstring{$r_{\text{semantic}}$}{r\_semantic}}
\label{app:prompt-semantic}

Both training and evaluation use Kimi-K2.5 for outcome semantic judgment with the following prompt. The output is strict JSON containing \texttt{intent\_consistency}, \texttt{confidence\_score}, and \texttt{detailed\_analysis}.

\begin{verbatim}
Please analyze the semantic consistency between the code repair generated by the
model and the ground-truth (GT) repair.

**Compilation error information:**
{error_info}

**Model-generated repair:**
{model_patches_str}

**GT repair:**
{gt_patches_str}

**Semantic-consistency evaluation criteria:**
1. **Semantic equivalence**: The repair generated by the model is semantically
   equivalent to the GT repair.
2. **Functional consistency**: The repaired program produces the same output
   under the same input.
3. **Behavior preservation**: The repair does not change the logical flow or
   computation process of the program.

**Evaluation requirements:**
- Compare only the modifications related to the target compilation error.
- The GT may contain changes unrelated to the target error; these
  should not affect the judgment.
- Focus on the semantic intent and runtime behavior of the repair.

**Return the result strictly in the following JSON format, with no extra text:**

{
    "is_semantic_consistent": true,
    "intent_consistency": "consistent",
    "confidence_score": 0.8,
    "detailed_analysis": "Detailed analysis and rationale",
    "key_differences": ["difference 1", "difference 2"]
}
\end{verbatim}

\subsection{Process-Level Semantic Scoring Prompt: Step-Level Weights in GRPO-full}
\label{app:prompt-process}

GRPO-full uses Kimi-K2.5 to score the complete tool trajectory step by step, producing a \texttt{score} $\in [0,1]$ and a brief \texttt{reason} for each step. These scores are used to reweight the loss mask by step.

\begin{verbatim}
You are a reward model for code repair. Please evaluate the directional
correctness and information gain of the entire tool trajectory based on semantic
reasoning.

[Compilation error information]
{error_info}

[Model tool trajectory, step by step]
{tool_calls_str}

[Ground-truth (GT) repair used for semantic comparison]
{gt_patches_str}

Semantic-consistency evaluation criteria. The comparison with GT should use
semantic judgment rather than exact matching:
1. Semantic equivalence: a model-generated repair is correct if it is
   semantically equivalent to the GT.
2. Functional consistency: the repaired program should produce the same output
   under the same input.
3. Behavior preservation: the repair should not change the logical flow or
   computation process of the program.

Evaluation requirements:
- Compare only the modifications related to the target compilation error.
- The GT may contain modifications unrelated to the target error; these
  should not affect the judgment.
- Focus on the semantic intent and runtime behavior of the repair.
- Repeatedly viewing files or regions that have already been inspected should be
  penalized as low information gain.
- compile_code is used to verify the repair and is part of the normal workflow;
  it should receive a relatively high score.

Output JSON only, with no extra text:
{
  "process_score": 0.0,
  "step_process_scores": [
    {"step_index": 0, "score": 0.0, "reason": "Brief reason"}
  ],
  "global_comment": "Overall process assessment"
}
\end{verbatim}

\subsection{Hint-Type Construction and Selection}
\label{app:hint-design}

\paragraph{Design motivation.}
The GT for the compile-fix task is the final code state, represented as a diff, rather than an action sequence, and the compiler already provides rich error-localization information. Our privileged information is structurally extracted from GT diffs in the annotated data and requires no additional inference cost. Data analysis shows that nearly half of the repairs occur in files not directly reported by the compiler, while error logs contain many noisy files. File-localization information therefore provides substantial incremental value.

\paragraph{Three basic hint types.}

Table~\ref{tab:hint-types} summarizes the construction form, information density, and intended use of the three basic hint types. All types are generated from the same metadata, including GT patches, error types, and error logs, and differ only in their level of abstraction.

\begin{table}[t]
    \centering
    \scriptsize
    \setlength{\tabcolsep}{4pt}
    \renewcommand{\arraystretch}{1.05}
    \caption{Construction forms and information density of the three basic hint types.}
    \label{tab:hint-types}
    \begin{tabular}{@{}p{0.12\linewidth} p{0.22\linewidth} p{0.17\linewidth} p{0.41\linewidth}@{}}
        \toprule
        \textbf{Type} & \textbf{Construction form} & \textbf{Information density} & \textbf{Design intent} \\
        \midrule
        Type A\newline (GT Diff)      & Inject the GT diff as a teacher-side upper-bound hint in the system prompt & Highest: upper-bound hint & Validate the performance upper bound; excluded from no-hint evaluation \\
        \addlinespace[2pt]
        Type B\newline (File localization) & Extract file paths to modify, line ranges, error-type categories, and repair directions & Medium: WHERE & Help localization without directly giving the answer, balancing information content and generalization \\
        \addlinespace[2pt]
Type C\newline (NL strategy)  & Use glm-5 to abstract an NL repair strategy of approximately $150$ words from the GT diff & Low: HOW & Minimize KL cost while encouraging error-analysis ability \\
        \bottomrule
    \end{tabular}
\end{table}

\paragraph{Type A example: upper-bound hint.}
\begin{verbatim}
## Teacher-side upper-bound hint
The following information is used only to construct a hinted teacher for
diagnostic upper-bound analysis and is not used in no-hint evaluation:

### Modification 1: condition_matcher.h
```diff
@@ -108,6 +108,7 @@ class ConditionMatcher {
      MMERR("cond.values.size < 1");
      return result;
    }
+
    switch (field.type) {
      case FieldType::UINT64: {
@@ -153,6 +154,9 @@
        break;
      }
+      default:
+        MMERR("invalid field type");
+        break;
    }
```
\end{verbatim}

\paragraph{Type B example: file localization.}
\begin{verbatim}
## Repair hint
- Error type: warning_as_error
- Files that need modification:
  - compute/matcher/condition_matcher.h (around line 111)
- Repair direction: eliminate the compilation warning, which has been promoted
  to an error by -Werror.
\end{verbatim}

\paragraph{Type C example: NL strategy.}
\begin{verbatim}
## Repair strategy hint
This compilation error occurs because the switch statement does not handle all
enumeration values. The compiler enables -Werror=switch, which promotes this
warning to an error. The repair is to add a default branch at the end of the
switch statement to handle unlisted enum values.
\end{verbatim}

\paragraph{Multi-level NL hint granularity experiments: C1--C5.}

Based on the Phase~0 result that Type~C is safe and effective, we further design five granularity levels. All are produced by LLM calls that abstract a general repair strategy from the GT diff, and they strictly avoid leaking any file names, function names, variable names, line numbers, or code snippets. Table~\ref{tab:hint-c-levels} shows the construction goal and abstraction level of each hint level.

\begin{table}[t]
    \centering
    \scriptsize
    \setlength{\tabcolsep}{4pt}
    \renewcommand{\arraystretch}{1.05}
    \caption{Construction goals and abstraction levels of five NL hint granularities.}
    \label{tab:hint-c-levels}
    \begin{tabular}{@{}c p{0.22\linewidth} c p{0.46\linewidth}@{}}
        \toprule
        \textbf{Level} & \textbf{Construction goal} & \textbf{Target length} & \textbf{Prompt guidance} \\
        \midrule
C1 & Error-pattern recognition & $\sim\!60$ words  & One-sentence summary of the error pattern and general repair direction \\
C2 & Repair-strategy direction & $\sim\!120$ words & General diagnostic procedure and repair-strategy direction \\
C3 & Repair reasoning chain     & $\sim\!180$ words & Reasoning chain from observing the error to selecting the repair, including key judgment points \\
C4 & Step-by-step methodology   & $\sim\!300$ words & A 3--5 step general methodology and judgment principle for each step \\
C5 & Complete repair guide      & $\sim\!500$ words & Complete guide covering diagnosis, strategy, verification, and pitfalls \\
        \bottomrule
    \end{tabular}
\end{table}

\paragraph{C1--C5 inference results.}

We inject hints during inference on $400$ evaluation samples at the step-$239$ checkpoint, using temperature $0.7$ and pass@1. The results are shown in Table~\ref{tab:hint-c-results}. C3, an NL strategy summary of approximately $180$ words, obtains the highest pass@1 of $53.5\%$, improving over the no-hint baseline of $44.5\%$ by $+9.0$pp. C1 provides only $+2.5$pp because it contains too little information, while C4 and C5 cause more fallback cases because of information overload, yielding lower net gains than C3.

\begin{table}[t]
    \centering
    \scriptsize
    \setlength{\tabcolsep}{4pt}
    \renewcommand{\arraystretch}{1.05}
    \caption{Inference results for five NL hint granularities at step $239$ on $400$ samples.}
    \label{tab:hint-c-results}
    \resizebox{\linewidth}{!}{%
    \begin{tabular}{@{}l r r r r r@{}}
        \toprule
        \textbf{Hint type} & \textbf{Compile success} & \textbf{Semantic consistency} & \textbf{pass@1} & \textbf{$\Delta$ pass@1} & \textbf{Net gain} \\
        \midrule
Baseline (no hint)             & $88.8\%$          & $45.5\%$          & $44.5\%$          & ---               & ---            \\
        \addlinespace[1.5pt]
C1 ($\sim\!60$ words)          & $91.8\%$          & $48.0\%$          & $47.0\%$          & $+2.5$            & $10$           \\
C2 ($\sim\!120$ words)         & $91.2\%$          & $50.5\%$          & $49.5\%$          & $+5.0$            & $20$           \\
\textbf{C3 ($\sim\!180$ words)} & $\mathbf{92.2\%}$ & $\mathbf{53.5\%}$ & $\mathbf{53.5\%}$ & $\mathbf{+9.0}$   & $\mathbf{36}$  \\
C4 ($\sim\!300$ words)         & $88.8\%$          & $50.0\%$          & $48.2\%$          & $+3.7$            & $15$           \\
C5 ($\sim\!500$ words)         & $90.0\%$          & $50.5\%$          & $49.8\%$          & $+5.2$            & $21$           \\
        \bottomrule
    \end{tabular}%
    }
\end{table}

\paragraph{Rationale for choosing C3.}
The experiments show a clear inverted-U curve: C1 is under-informative; C2 provides the right direction but lacks a reasoning chain; C3 strikes the best balance among error-pattern recognition, diagnostic procedure, and key judgment criteria; and C4/C5 are too long, causing attention dispersion and more fallback cases, with $41$ fallback cases for C4 versus $34$ for C3. In addition, C3 has the highest \texttt{str\_replace} success rate among the five levels, $87.5\%$, and a slightly lower average number of steps, $17.9$, than the baseline, $19.1$, indicating that it guides the repair direction without adding extra operational burden. Therefore, the teachers in both $\pi$-Distill and OPSD use the C3 hint type.

\subsection{C3 Repair-Strategy Hint Generation Prompt}
\label{app:prompt-hint-gen}

The teachers in both $\pi$-Distill and OPSD use C3 hints, approximately $180$ words. The following prompt calls glm-5 to abstract a general repair reasoning chain from the GT diff, requiring that no concrete file names, function names, or code snippets be included.

\begin{verbatim}
You are an expert in fixing C++ compilation errors. Below is a compilation error
and the corresponding correct repair patch.

Your task: abstract from this patch the reasoning process for fixing this type
of error, describing the chain of thought from "seeing the error" to "choosing
the repair".

Strict requirements:
- Describe a general reasoning process: what information to inspect first, what
  judgment to make, what conclusion to draw, and what repair strategy to adopt.
- Focus on key judgment points and decision criteria, such as: "when an error
  reports an undefined symbol, determine whether the declaration is missing or
  whether the declaration is in the wrong scope".
- The strategy must be general and transferable to all errors of the same type.
- Do not include any file names, function names, variable names, line numbers, or
  concrete code snippets.
- Keep the total length strictly within 180 words.

Compilation error type: {error_types}
Compilation error log:
{errors}

Repair patch:
{diffs}
\end{verbatim}

\subsection{Privileged-Hint Injection Format}
\label{app:prompt-hint}

$\pi$-Distill and OPSD wrap the teacher hint in \texttt{<Privileged hint>} tags and insert it at the beginning of the user message. This allows the teacher to receive strategy guidance before reading the compilation error log. The hint is used only for training-time distillation comparisons and hinted upper-bound analysis, and is strictly excluded from no-hint evaluation.

\begin{verbatim}
<Privileged hint>
The following is a teacher-side strategy hint abstracted from the annotated
repair.
{hint_text}
</Privileged hint>
\end{verbatim}

\newpage
\section*{NeurIPS Paper Checklist}

\begin{enumerate}

\item {\bf Claims}
    \item[] Question: Do the main claims made in the abstract and introduction accurately reflect the paper's contributions and scope?
    \item[] Answer: \answerYes{}
    \item[] Justification: The paper's main claim is limited to signal-reshaping conditions for weak-feedback agentic repair, and each claimed condition is tied to a corresponding controlled comparison: layered rewards are evaluated through the GRPO-early versus GRPO-binary contrast, process-level credit through the GRPO-early versus GRPO-full contrast, and token-level distillation is used as a boundary comparison. This keeps the claims aligned with the evidence actually reported.

\item {\bf Limitations}
    \item[] Question: Does the paper discuss the limitations of the work performed by the authors?
    \item[] Answer: \answerYes{}
    \item[] Justification: Several factors could limit generalization, including the use of compile-fix as the representative task, reliance on private CI-derived repair data, a fixed 400-sample evaluation split, and single-run RL comparisons. The manuscript narrows its conclusions to this weak-feedback repair setting and uses the distillation failure analysis to show that not all forms of privileged supervision transfer to this regime.

\item {\bf Theory assumptions and proofs}
    \item[] Question: For each theoretical result, does the paper provide the full set of assumptions and a complete (and correct) proof?
    \item[] Answer: \answerNA{}
    \item[] Justification: The formal equations are used to define the MDP, masking, GRPO objective, and distillation baselines so that the experiments are unambiguous. They are not presented as theorems with novel guarantees, so proof obligations do not arise.

    \item {\bf Experimental result reproducibility}
    \item[] Question: Does the paper fully disclose all the information needed to reproduce the main experimental results of the paper to the extent that it affects the main claims and/or conclusions of the paper (regardless of whether the code and data are provided or not)?
    \item[] Answer: \answerNo{}
    \item[] Justification: Independent reproduction depends on private CI-derived examples, repository revisions, build scripts, and sandbox infrastructure that are not publicly available. The manuscript therefore exposes the reward mapping, rollout controls, prompts, and hyperparameters for auditability, but cannot by itself provide all inputs needed to rerun the experiments end to end.

\item {\bf Open access to data and code}
    \item[] Question: Does the paper provide open access to the data and code, with sufficient instructions to faithfully reproduce the main experimental results, as described in supplemental material?
    \item[] Answer: \answerNo{}
    \item[] Justification: The data and execution traces come from private CI-derived repair records and depend on repository snapshots, build infrastructure, and governance constraints that cannot be redistributed in anonymized form at submission time. To make the method auditable despite this constraint, the manuscript documents the task construction, reward logic, rollout governance, and prompt templates in the appendix.

\item {\bf Experimental setting/details}
    \item[] Question: Does the paper specify all the training and test details (e.g., data splits, hyperparameters, how they were chosen, type of optimizer) necessary to understand the results?
    \item[] Answer: \answerYes{}
    \item[] Justification: Understanding the comparisons requires knowing which components are held fixed and which training signal changes across GRPO-early, GRPO-binary, and GRPO-full. The manuscript gives those controls together with the base model, dataset sizes, fixed evaluation split, optimizer settings, rollout budget, sampling parameters, and reward construction.

\item {\bf Experiment statistical significance}
    \item[] Question: Does the paper report error bars suitably and correctly defined or other appropriate information about the statistical significance of the experiments?
    \item[] Answer: \answerNo{}
    \item[] Justification: The reported runs require expensive agentic rollouts with Docker execution and compilation, and the current study prioritizes controlled mechanism comparisons over repeated-seed estimation. Appendix~\ref{app:judge-validation} reports a paired McNemar test and confidence interval for the main post-fork GRPO-full versus GRPO-early comparison, but the broader training curves are single-run fixed-split comparisons without repeated-seed error bars.

\item {\bf Experiments compute resources}
    \item[] Question: For each experiment, does the paper provide sufficient information on the computer resources (type of compute workers, memory, time of execution) needed to reproduce the experiments?
    \item[] Answer: \answerNo{}
    \item[] Justification: Appendix~\ref{app:limitations} now describes the main compute bottlenecks, shared sampling budget, three-way resource-control policy for GPU inference, sandbox I/O, and compilation, and wall-clock spans for the four GRPO comparison runs. Exact total project compute is still omitted because the runs were conducted on a shared internal cluster with variable queueing and background load.
    
\item {\bf Code of ethics}
    \item[] Question: Does the research conducted in the paper conform, in every respect, with the NeurIPS Code of Ethics \url{https://neurips.cc/public/EthicsGuidelines}?
    \item[] Answer: \answerYes{}
    \item[] Justification: The study uses private CI-derived failure records and developer-submitted repairs rather than collecting new human-subject data, and the manuscript anonymizes project-specific paths and infrastructure identifiers. The experiments are confined to sandboxed repair environments and do not deploy trained agents to production systems.

\item {\bf Broader impacts}
    \item[] Question: Does the paper discuss both potential positive societal impacts and negative societal impacts of the work performed?
    \item[] Answer: \answerYes{}
    \item[] Justification: Appendix~\ref{app:limitations} now discusses both the potential maintenance benefit of weak-feedback code-repair agents and deployment risks from agents that lack semantic validation and signal governance, including behavior-changing repairs and unsafe autonomous use without sandboxing, provenance tracking, or human review.
    
\item {\bf Safeguards}
    \item[] Question: Does the paper describe safeguards that have been put in place for responsible release of data or models that have a high risk for misuse (e.g., pre-trained language models, image generators, or scraped datasets)?
    \item[] Answer: \answerNA{}
    \item[] Justification: The work reports analysis and aggregate experimental results, while the underlying CI-derived data, trained checkpoints, and execution environment are not released with the submission. Since no deployable model or dataset is released, release-specific safeguards such as gated access or usage restrictions are not applicable.

\item {\bf Licenses for existing assets}
    \item[] Question: Are the creators or original owners of assets (e.g., code, data, models), used in the paper, properly credited and are the license and terms of use explicitly mentioned and properly respected?
    \item[] Answer: \answerYes{}
    \item[] Justification: The external models and methods are used as cited baselines, policies, or conceptual references rather than redistributed assets, and they are identified in the main text and bibliography. The remaining data source is private CI-derived material governed by access controls, so no third-party dataset is repackaged or relicensed.

\item {\bf New assets}
    \item[] Question: Are new assets introduced in the paper well documented and is the documentation provided alongside the assets?
    \item[] Answer: \answerNA{}
    \item[] Justification: The paper introduces an experimental setup and reports prompt templates for reproducibility context, but it does not package these materials as a public dataset, benchmark, model, or software artifact. Therefore there is no standalone asset card, license, or distribution documentation to provide.

\item {\bf Crowdsourcing and research with human subjects}
    \item[] Question: For crowdsourcing experiments and research with human subjects, does the paper include the full text of instructions given to participants and screenshots, if applicable, as well as details about compensation (if any)? 
    \item[] Answer: \answerNA{}
    \item[] Justification: The study analyzes existing compilation failures and code repairs produced during normal development workflows. It does not recruit participants, assign tasks to crowd workers, collect screenshots, or provide experimental instructions to human subjects.

\item {\bf Institutional review board (IRB) approvals or equivalent for research with human subjects}
    \item[] Question: Does the paper describe potential risks incurred by study participants, whether such risks were disclosed to the subjects, and whether Institutional Review Board (IRB) approvals (or an equivalent approval/review based on the requirements of your country or institution) were obtained?
    \item[] Answer: \answerNA{}
    \item[] Justification: No intervention, survey, user study, or human-subject data collection is performed. The analyzed artifacts are CI logs and code diffs handled under engineering-data governance, so there are no study participants whose experimental risk or consent process would need IRB reporting.

\item {\bf Declaration of LLM usage}
    \item[] Question: Does the paper describe the usage of LLMs if it is an important, original, or non-standard component of the core methods in this research? Note that if the LLM is used only for writing, editing, or formatting purposes and does \emph{not} impact the core methodology, scientific rigor, or originality of the research, declaration is not required.
    %this research? 
    \item[] Answer: \answerYes{}
    \item[] Justification: LLMs are part of the experimental object and measurement pipeline: the repair policy is an LLM agent, semantic correctness is judged by an LLM, process scores are assigned by an LLM, and privileged-hint teachers are LLM variants. Because these uses affect rewards, training signals, and conclusions, the manuscript specifies their roles and prompt templates.

\end{enumerate}

\end{document}